\pdfoutput=1

\documentclass[11pt]{article}

\usepackage{acl}

\usepackage{times}
\usepackage{latexsym}
\usepackage{graphicx}
\usepackage{multirow}
\usepackage{comment}
\usepackage{caption}
\usepackage{longtable}

\usepackage[T1]{fontenc}

\usepackage[utf8]{inputenc}

\usepackage{microtype}
\usepackage{booktabs}
\usepackage{xspace}

\usepackage{amssymb} 
\usepackage{pifont} 

\newcommand*{\sib}{\textsc{SIB-200 \xspace}}

\newcommand{\cmark}{\ding{51}}

\title{SIB-200: A Simple, Inclusive, and Big Evaluation Dataset for Topic Classification in 200+ Languages and Dialects}

\author{David Ifeoluwa Adelani$^{1,2*}$, \ Hannah Liu$^{3}$\thanks{\ \ Equal contribution and corresponding authors}, \ Xiaoyu Shen$^{4*}$\thanks{\ \ Work done outside Amazon}, \ Nikita Vassilyev$^{3}$,  \ \\ \textbf{Jesujoba O. Alabi$^{2,5}$, \  Yanke Mao$^{3}$, \ Haonan Gao$^{3}$, and En-Shiun Annie Lee$^{3,6}$.}\\
$^{1}$University College London, $^{2}$Masakhane, $^{3}$University of Toronto,  $^{4}$Amazon Alexa AI, \\
$^{5}$Saarland University, $^{6}$University of Ontario Institute of Technology\\
\texttt{d.adelani@ucl.ac.uk, hannahhere.liu@mail.utoronto.ca, } \\
\texttt{ gyouu@amazon.com} \\
}

\begin{document}

\newcommand{\da}[1]{{\textcolor{orange}{[David: #1]}}}
\newcommand{\eal}[1]{{\textcolor{cyan}{[#1]}}}
\newcommand{\as}[1]{{\textcolor{green}{[Aaron: #1]}}}
\newcommand{\nik}[1]{{\textcolor{red}{[Nikita: #1]}}}
\newcommand{\ym}[1]{{\textcolor{blue}{[Yanke: #1]}}}

\newcommand{\sh}[1]{{\textcolor{teal}{[Ray: #1]}}}

\maketitle
\begin{abstract}

Despite the progress in building multilingual language models, evaluation is often limited to a few languages with available datasets which excludes a large number of low-resource languages. 
In this paper, we create \textsc{SIB-200}---a large-scale open-sourced benchmark dataset for topic classification in 205 languages and dialects to address the lack of evaluation dataset for Natural Language Understanding (NLU). For many of the languages covered in \textsc{SIB-200}, this is the first publicly available evaluation dataset for NLU. The dataset is based on Flores-200 machine translation corpus. We annotated the English portion of the dataset and extended the sentence-level annotation to the remaining 204 languages covered in the corpus. Despite the simplicity of this task, our evaluation in full-supervised setting, cross-lingual transfer setting and prompting of large language model setting show that there is still a large gap between the performance of high-resource and low-resource languages when multilingual evaluation is scaled to 
numerous world languages. We found that languages unseen during the pre-training of multilingual language models, languages from under-represented families (like Nilotic and Altantic-Congo), and languages from the regions of Africa, Americas, Oceania and South East Asia, often have the lowest performance on our topic classification dataset. We hope our dataset 
encourage a more inclusive evaluation of multilingual language models on a more diverse set of languages.~\footnote{\url{https://github.com/dadelani/SIB-200}}
\end{abstract}

\section{Introduction}

In the last few years, developing massively multilingual Pre-trained Language Models (PLMs) to scale to several written languages is an active area of research---e.g. covering 100 languages~\cite{devlin-etal-2019-bert,conneau-etal-2020-unsupervised,liu-etal-2020-multilingual-denoising,xue-etal-2021-mt5,he2023debertav}. 
However, evaluation is often limited to a few tens of languages with benchmark datasets~\citep{conneau-etal-2018-xnli,hu2020xtreme,ruder-etal-2021-xtreme,zhang2022mdia}, 
thus, limiting the large-scale evaluation of current multilingual language models on many languages, especially those 
truly low-resource languages. 

While there is evidence from previous works that languages not covered during pre-training often lead to lower performance, such analysis is also limited to a small selection of languages with annotated datasets~\cite{ponti-etal-2020-xcopa,pfeiffer-etal-2020-mad,adelani-etal-2022-masakhaner, lee-etal-2022-pre}. 

Recently, there is a push to scale evaluation datasets to more than 100 languages, but this requires a very expensive annotation effort in terms of money and time. Often, this scaling is only carried out by a large community effort that spans many years like the Universal Dependency (UD) project ~\citep{nivre-etal-2017-universal,nivre-etal-2020-universal,de-marneffe-etal-2021-universal} or financed by BigTech companies~\citep{goyal-etal-2022-flores,team2022NoLL,federmann-etal-2022-ntrex,fleurs,Pratap2023ScalingST}. 
Despite these investments in data curation, there are only few benchmarks for natural language understanding (NLU) tasks that cover all the languages seen during the pre-training of multilingual PLMs~\citep{imanigooghari-etal-2023-glot500}. 

The largest benchmark datasets that are available for NLU are UD, Taxi1500~\citep{Ma2023Taxi1500AM}, WikiANN~\cite{pan-etal-2017-cross}, and Belebele~\citep{bandarkar2023belebele} for dependency parsing, text classification, named entity recognition, and reading comprehension,  respectively. The largest is Taxi-1500 for 1500 languages---but it is biased to the religious domain, and some languages are not publicly available due to copyright. WikiANN on the other hand, was automatically annotated and with few instances for low-resource languages. UD and Belebele were manually annotated and covered between 100-125 languages. However, many languages are still missing in the above datasets. 

In this paper, we create \textsc{SIB-200}---a large-scale open-sourced benchmark dataset for topic classification to address the lack of evaluation datasets for NLU. The dataset is based on Flores-200~\citep{team2022NoLL}---a multi-way parallel corpus (i.e. same sentences are available in 205 languages). We annotated the English portion of the Flores-200 dataset and extend the sentence-level annotation to the remaining 204 languages covered in Flores-200. 

Our evaluation shows that there is still a large gap between the performance of high-resource and low-resource languages when multilingual evaluation is scaled to
numerous world languages. Languages unseen during the pre-training of multilingual PLMs, languages from under-represented families (like Nilotic and Altantic-Congo), and languages from the regions of Africa, Americas, Oceania and South East Asia, often have the lowest performance on our 
dataset. We also find that simply scaling up the number of languages without scaling up the domains in the pre-training is unhelpful (e.g., Glot-500 pre-trained on 500 languages largely under-performs XLM-R pre-trained on 100 languages). It is crucial to mix text from various domains. For languages unseen during pre-training, we show the potential of multilingual language adaptive fine-tuning (MAFT)\footnote{adaptation of an existing multilingual PLM to multiple or new sets of languages simultaneously.}~\citep{Tang2020MultilingualTW, alabi-etal-2022-adapting} in improving the performance of these languages by leveraging synthetic data for languages with tiny monolingual data (i.e. language with less than 10MB of data). Evaluation of this approach on African languages results in significant improvement (up to $+5\%$ in accuracy on average) for the previously unseen languages. 

Finally, we extend our evaluation to the zero-shot settings by training individually on English, French, Arabic and Chinese (Simplified) languages using XLM-R~\citep{conneau-etal-2020-unsupervised}, and performing zero-shot evaluation on other languages. We compared these results with prompting GPT-4 large language models (LLMs). Our results show that LLMs perform poorly on over 64.7\% (or 132 out of 205 languages) of the languages with less than 70\% in accuracy while zero-shot adaptation from the English model only leads to performance less than 70\% accuracy in 81 languages (or 39.3\% of languages)\footnote{Performance of XLM-R on English is 92.1\% in accuracy while prompting GPT-4 in English gave 76.6\% in accuracy.}. This shows that leveraging cross-lingual transfer from high-resource languages is much better than prompting LLMs for many languages. 


\section{\sib dataset}

\begin{table}[t] 
\centering
\resizebox{\columnwidth}{!}{%
\begin{tabular}{lrrrr}
\toprule
\textbf{Label} & \textbf{TRAIN} & \textbf{DEV} & \textbf{TEST} & \textbf{TOTAL}  \\ 
\midrule
science/technology  & 176 & 25 & 51 & 252 \\ 
travel  & 138 & 20 & 40 & 198 \\ 
politics  & 102 & 14 & 30 & 146 \\ 
sports  & 85 & 12 & 25 & 122 \\ 
health  & 77 & 11 & 22 & 110 \\ 
entertainment  & 65 & 9 & 19 & 93 \\ 
geography  & 58 & 8 & 17 & 83 \\ 
\midrule
Total & 701 & 99 & 204 & 1,004\\
\bottomrule
\end{tabular}
}
\vspace{-2mm}
\caption{\textbf{\sib dataset}. We provide the data size of the annotated data by their SPLIT and category}
\label{tab:ftac_dataset}

\end{table}

\subsection{Data source}
\label{sec:data_collection}
We introduce 
\textsc{SIB-200}---a \underline{\textbf{S}}imple \underline{\textbf{I}}nclusive and \underline{\textbf{B}}ig topic classification 
dataset for over 200 languages and dialects. We leveraged the multi-way parallel Flores-200 dataset~\citep{team2022NoLL} for the creation of the dataset. Flores-200 corpus is an extension of Flores-101~\citep{goyal-etal-2022-flores}---for 101 languages. In both datasets, the source sentences were collected in English and translated by professional translators to several languages. In total, the corpus contains 3,001 sentences divided into \texttt{DEV} (997 sentences), \texttt{DEVTEST} (1,012 sentences) and \texttt{TEST} (992 sentences) sets. However, the authors did not release the \texttt{TEST} set. Additionally, we added N’Ko---a West African language that was recently added to Flores-200 dataset~\citep{doumbouya-etal-2023-machine}.~\footnote{\url{https://oldi.org/}} 

Flores-200 released additional information to provide meta-data information about the domains and topics of the articles covered in the dataset. The domains are based on WikiNews, WikiJunior, and WikiVoyage with a total of 842 articles while the topics are based on ``crime'', ``disasters'', ``entertainment'', ``geography'', ``health'', ``nature'', ``politics'', ``science'', ``sports'', and ``travel''.~\footnote{We note that in the open-sourced dataset, there are more categories than the ten reported in the paper.} However, a quick review of the dataset revealed that at the sentence level, the article can belong to more than one topic. Therefore, we decided to add our topic categorization at the sentence level. Performing annotation at the sentence level also gives us the additional advantage of having more samples to annotate (2,009 rather than 562 instances\footnote{Although 842 articles are in Flores-200, only 562 articles are open-sourced as part of DEV and DEVTEST sets.}).

\subsection{Data annotation}
We recruited four annotators who are native speakers of English to label 2,009 sentences obtained from the \texttt{DEV} and \texttt{DEVTEST} sets of Flores-200\footnote{All annotators are also authors of this paper.}. We make use of an internal annotation tool for text classification. The annotation labelling scheme covers 15 categories, 10 are from the original Flores-200 categorization of articles (\S\ref{sec:data_collection}), and the others are ``business'', ``religion'', ``technology'', ``education'', and ``uncategorized''. We assigned sentences that do not fit any of the defined categories, and sentences lacking sufficient context about their topic to ``uncategorized''. An example of a sentence labelled as ``uncategorized'' is \textit{``In Berlin, police estimated 6,500 protestors''}. 
The annotators took about two weeks to complete the task, however on average it takes up to 60 seconds to annotate a sentence (approximately, 33 hours of annotation time). 


\subsection{Quality control}
We report Fleiss Kappa score~\citep{fleiss1971mns} to measure the agreement of annotation. The Fleiss Kappa score among the four annotators is \textbf{0.44}---which signifies a moderate level of agreement. 

\paragraph{Choosing the final label per sentence} We assigned the final label to a sentence by majority voting. Specifically, we assign a label to a sentence if at least two annotators agree on the category, but we excluded the situation, where any two annotators conflicted with the other two annotators. For example, for the sentence \textit{``The major organ of the circulatory system is the heart, which pumps the blood.''}, the first two annotators assigned ``science'' while the last two assigned ``health''. In total, we assigned a single label to 1,695 sentences, but there were 314 sentences with conflicts in the annotation. We asked the lead annotator to adjudicate the sentences with conflicting annotations and assigned a single label to each sentence. We later combined the fixed conflicting annotations with the others to give us back a total of 2009 annotated sentences.

\paragraph{Final classification dataset} 

\begin{table*}[t]
    \centering
    \resizebox{\textwidth}{!}{%
    \begin{tabular}{l|rrrrrrr|rr|p{11.7cm}}
        \toprule
         & \multicolumn{7}{c}{\textbf{Joshi's class}} & \multicolumn{2}{c}{\textbf{NLLB class}} & \\ 
        \textbf{Region} & \textbf{0} & \textbf{1} & \textbf{2} & \textbf{3} & \textbf{4} & \textbf{5} & \textbf{None} & \textbf{LRL} & \textbf{HRL} & \textbf{Language Families} \\ 
        \midrule
        Africa & 10 & 20 & 9 & 2 & -- & -- & 16 & 51 & 6 & Atlantic-Congo (34), Afro-Asiatic (12), Nilotic (5), Indo-European (2), Mande (2), Austronesian (1) \\ 
        Americas & 1 & 3 & -- & -- & -- & -- & 1 & 5 & -- & Indo-European (2), Aymaran (1), Tupian (1), Quechuan (1) \\ 
        Asia 1 (W \& C) & 2 & 8 & -- & 4 & 2 & 1 & 7 & 18 & 6 & Afro-Asiatic (8), Turkic (8), Indo-European (7), Kartvelian (1) \\ 
        Asia 2 (S) & 4 & 14 & 2 & 3 & 1 & -- & 3 & 25 & 2 & Indo-European (19), Dravidian (4), Sino-Tibetan (3), Austroasiatic (1) \\ 
        Asia 3 (SE \& E) & 3 & 17 & 1 & 5 & 2 & 2 & 1 & 22 & 9 & Austronesian (17), Sino-Tibetan (6), Tai-Kadai (3), Austroasiatic (2), Japonic (1), Mongolic-Khitan (1),  Koreanic (1) \\ 
        Europe 1 (N, W, S) & 1 & 17 & 3 & 7 & 10 & 4 & -- & 19 & 23 & Indo-European (36), Uralic (3), Constructed (1), Basque (1), Afro-Asiatic (1) \\ 
        Europe 2 (E) & -- & 6 & -- & 6 & 3 & -- & -- & 7 & 8 & Indo-European (12), Turkic (3) \\ 
        Oceania & -- & 4 & -- & -- & -- & -- & -- & 4 & -- & Austronesian (3), Indo-European (1) \\ 
        \midrule
        \textbf{Total} & 21 & 90 & 17 & 30 & 22 & 12 & 28 & 151 & 54 &  \\ 
        \bottomrule
    \end{tabular}
    }
    \vspace{-2mm}
    \caption{Language families covered in \textbf{\sib dataset} grouped by United Nations geoscheme \textbf{regions}, \textbf{Joshi's classes}~\citep{joshi-etal-2020-state} (None -- for languages not found in Joshi's dataset), and \textbf{NLLB classification}~\citep{team2022NoLL} of languages by the size of resources on the internet---High-resource language (HRL) or low-resource language (LRL). }
\label{tab:languages}
\end{table*}

For the final dataset, we excluded sentences with the label of ``uncategorized'', we only selected label categories with more than 80 sentences, this removed categories such as ``business'' (80 sentences), ``disasters'' (73 sentences), ``crime'' (72 sentences), ``education'' (52 sentences), and ``religion'' (46 sentences). We note that having too many categories with few sentences makes building text classification models a bit difficult leading to a lower performance. Also, we combined ``science'' (138 sentences) and ``technology'' (114 sentences) category into a single category of ``science/technology''. Finally, we removed the ``nature'' category because there is a lot of conflict with ``science'' and ``geography'' categories. Our preliminary experiments show that adding ``nature'' significantly lowers the performance of our classifier. About half of the Flores-200 is part of the \sib 
 dataset (i.e. 1004 out of 2009 sentences).

\autoref{tab:ftac_dataset} shows the number of sentences per label in each of the \texttt{TRAIN}, \texttt{DEV}, and \texttt{TEST} splits. We divided the sentences into the split using the 70\%, 10\%, 20\% ratio. The dataset will be released under CC BY-SA 4.0 licence. While the SIB-200 dataset only includes seven labels, we are also releasing another version of the dataset that is more challenging with all the 14 labels (excluding ``uncategorized''). We compared the performance of English dataset using both seven and 14 labels in \autoref{appendix:eng_result}.

\section{Experimental setup}

\begin{table*}[t]
    \centering
    \scalebox{0.775}{
    \begin{tabular}{lr|cccc|cccc|cc}
        \toprule
        \textbf{} &  & \multicolumn{4}{c|}{\textbf{Fully Supervised}} &\multicolumn{4}{c|}{\textbf{Cross-Lingual Transfer (XLMR)}} & \multicolumn{2}{c}{\textbf{Zero-Shot Prompt}}\\\textbf{Language Family} & \textbf{\small{Count}}&\textbf{\small{MLP}} & \textbf{\small{Glot-500}} & \textbf{\small{XLM-R (base)}} & \textbf{\small{XLM-R}} & 
        \textbf{\small{English}} & \textbf{\small{French}} & \textbf{\small{Chinese}} & \textbf{\small{Arabic}} & \textbf{\small{GPT-3.5-Turbo}} & \textbf{\small{GPT-4}} \\ 
        \midrule
        English & - & 59.9 & 82.8 & 90.0 & 92.1 &92.1&91.9&\textbf{92.5}&91.2&71.8&76.6\\ 
        \midrule
        Indo-European & 79 & 62.3 & 72.4 & 81.4 & \textbf{86.2} & 82.4 & 83.2 & 82.8 & 83.0 & 55.3 & 66.6 \\ 
        Atlantic-Congo & 34 & \textbf{61.3} & 49.6 & 50.5 & 57.9 & 41.4 & 41.4 & 41.9 & 42.0 & 29.2 & 29.2 \\ 
        Afro-Asiatic & 21 & 61.4 & 59.2 & 67.1 & \textbf{72.6} & 67.4 & 68.1 & 67.7 & 68.4 & 43.4 & 54.6 \\ 
        Austronesian & 21 & 59.8 & 62.1 & 68.8 & \textbf{73.9} & 64.0 & 64.3 & 64.5 & 64.9 & 44.1 & 47.1 \\ 
        Turkic & 11 & 64.8 & 74.2 & 79.8 & \textbf{85.1} & 80.2 & 80.9 & 80.4 & 80.9 & 50.2 & 59.2  \\ 
        Sino-Tibetan & 9 & \textbf{68.8} & 66.2 & 62.2 & 65.4 & 57.9 & 58.3 & 57.1 & 57.1 & 30.7 & 40.6  \\ 
        Nilotic & 5 & \textbf{58.6} & 35.0 & 48.2 & 53.7 & 34.8 & 33.0 & 34.0 & 34.0 & 16.1 & 10.1  \\ 
        Dravidian & 4 & 64.7 & 76.1 & 84.4 & \textbf{87.9} & 87.8 & 88.1 & 88.2 & 88.0 & 57.2 & 69.6 \\ 
        Tai-Kadai & 3 & 67.7 & 61.3 & 70.9 & \textbf{76.8} & 68.4 & 67.8 & 68.9 & 69.2 & 35.6 & 44.7 \\ 
        Uralic & 3 & 62.1 & 74.1 & 86.5 & \textbf{89.6} & 89.1 & 90.4 & 90.2 & 89.6 & 62.4 & 74.8   \\ 
        Austroasiatic & 3 & 66.5 & 65.5 & 66.2 & \textbf{68.1} & 67.5 & 66.8 & 67.2 & 66.2 & 34.8 & 48.7 \\ 
        Mande & 2 & \textbf{57.4} & 36.1 & 42.7 & 48.7 & 32.5 & 32.4 & 32.3 & 32.1 & 18.0 & 13.3 \\ 
        Japonic & 1 & 73.8 & 81.5 & 87.9 & \textbf{89.9} & 89.3 & 90.3 & 89.7 & 88.8 & 63.4 & 75.8  \\ 
        Koreanic & 1 & 67.8 & 76.5 & 86.5 & \textbf{88.5} & 88.7 & 89.4 & 89.2 & 88.7 & 67.8 & 78.2 \\ 
        Mongolic-Khitan & 1 & 66.2 & 74.8 & 82.9 & \textbf{88.5} & 86.1 & 85.8 & 85.5 & 86.2 & 57.7 & 67.6 \\ 
        Constructed & 1 & 61.4 & 72.8 & 87.5 & \textbf{89.4} & 88.5 & 89.2 & 90.4 & 88.6 & 58.7 & 70.3  \\ 
        Quechuan & 1 & 53.7 & 59.4 & 57.9 & \textbf{64.1} & 46.3 & 48.3 & 49.1 & 50.8 & 36.2 & 18.5 \\ 
        Basque & 1 & 62.9 & 72.4 & 83.5 & \textbf{89.2} & 89.2 & 90.0 & 89.7 & 88.9 & 55.3 & 53.1 \\ 
        Aymaran & 1 & \textbf{55.7} & 37.4 & 42.5 & 52.5 & 39.1 & 40.4 & 38.5 & 41.3 & 15.9 & 6.6 \\ 
        Tupian & 1 & 57.7 & 63.7 & 69.6 & \textbf{76.3} & 61.3 & 61.7 & 61.7 & 61.1 & 32.3 & 28.2 \\ 
        Kartvelian & 1 & 63.7 & 78.4 & 83.4 & \textbf{88.5} & 89.1 & 89.8 & 89.7 & 88.6 & 44.7 & 66.1 \\
         \midrule
        Average & - & 62.8 & 64.2 & 71.0 & \textbf{75.9} & 69.1 & 69.5 & 69.5 & 69.5 & 43.3 & 48.7 \\
        \bottomrule
       
    \end{tabular}
    }
    \caption{{\bf Overall result of the performance of different text-classification models across different language families.} We compared different settings: fully-supervised, cross-lingual transfer and zero-shot prompting of LLMs. Cross-lingual transfer is based on the XLM-R model as it is the best-performing PLM. Performances from 4 source languages: English, French, Chinese and Arabic are reported.}
    \label{tab:baseline}
\end{table*}

Here, we describe different categorization of languages, text classification models developed for \sib, and the experimental settings (i.e. full supervised setting and zero-shot transfer setting).

\subsection{Languages and their categorizations}
\label{sec:plms_covered}

\autoref{tab:languages} and \autoref{tab:lang_family_plm} shows the grouping of languages in the \sib dataset. We categorized them based on the following characteristics:  
(1) geographical regions, (2) language family, (3) coverage in multilingual PLMs, and (4) Joshi's classification~\citep{joshi-etal-2020-state}---a  categorization based on their labelled/unlabelled resources on the web---making it easy to analyze results.

\paragraph{Categorization by geographical regions} \autoref{tab:languages} shows the grouping of languages into regions according to the United Nations Geoscheme\footnote{\url{https://en.wikipedia.org/wiki/United_Nations_geoscheme}}. The regions are: Africa, Americas, Asia 1 or Western \& Central Asia, Asia 2 or Southern Asia, Asia 3 or South-Eastern \& Eastern Asia, Europe 1 or Northern/Western/Southern Europe, Europe 2 or Eastern Europe, and Oceania. 

\paragraph{Categorization by language family} \sib languages are grouped into 21 language families as shown in \autoref{tab:lang_family_plm}, the largest groups are: Indo-European (79 languages), Atlantic-Congo (35 languages), Afro-Asiatic (21 languages), Austronesian (21 languages) and Turkic (11 languages).

\paragraph{Categorization by Joshi's classification} \autoref{tab:languages} also shows the number of languages in each Joshi's class---a measure of the unlabelled or labelled resources available for each language on the web~\citep{joshi-etal-2020-state}. 128 languages can be categorized as low-resource since they fall between class ``0'' and ``2'', 30 languages are mid-resource in class ``3'', and the others are high-resource (only 39 languages). This also corresponds to the NLLB classification for machine translation resources available on the web, but with only two categories---150 low-resource languages (LRLs) and 54 high-resource languages (HRLs). 

\paragraph{Categorization by availability in PLM} Lastly, we grouped languages and language families by their inclusion in the training of multilingual PLMs. XLM-R~\citep{conneau-etal-2020-unsupervised} covered 90 out of the 205 languages in our dataset while GLOT-500~\citep{imanigooghari-etal-2023-glot500} covered 177. This is a good indication of performance in general since languages that are included during pre-training often have better performance~\citep{ponti-etal-2020-xcopa,pfeiffer-etal-2020-mad,adelani-etal-2022-masakhaner}. 
Finally, we show the number of languages covered by region-specific PLMs such as AfriBERTa~\citep{ogueji-etal-2021-small}, AfroXLMR~\citep{alabi-etal-2022-adapting}, Serengeti~\citep{adebara-etal-2023-serengeti}, MuRIL~\citep{Khanuja2021MuRILMR}, and IndicBERTv2~\citep{doddapaneni-etal-2023-towards}. The grouping is provided in \autoref{tab:lang_family_plm}.

\subsection{Text classification models} 
\label{sec:text_class_models}
We trained a simple Multilayer Perceptron (MLP), fine-tuned multilingual PLMs and prompted large language models for text classification. 

\paragraph{Multi-Layer Perceptron}
For the input features, we make use of either n-gram features (n=1 up to 3 in our experiments) or XLM-R tokens obtained by first tokenizing the sentences using XLM-R tokenizer. We make use of the default setting on scikit-learn tool~\citep{scikit-learn}

\paragraph{Masked Language Models (MLM)}

Next, we fine-tune massively multilingual PLM such as XLM-R-base (270M parameters), XLM-R (550M) Glot-500 (395M), which are trained on several languages: XLM-R and Glot-500 were trained on 100 and 500 languages respectively. We also fine-tune region-specific PLM trained on multiple country-level or continent-level languages: AfriBERTa (126M), Serengeti (278M), AfroXLMR (550M), MuRIL (236M) and IndicBERTv2 (278M). We restrict region-level analysis to Africa and India because we only found these two regions with multilingual PLMs covering many languages.

\paragraph{MAFT with fewer data and synthetic data}

We explore how to improve over regional PLMs using MAFT---adaptation of an existing multilingual PLM to multiple or new set of languages simultaneously, this was effective for adapting XLM-R to 20 languages spoken in Africa~\citep{alabi-etal-2022-adapting}. To extend to more languages, we apply MAFT to 61 African languages with at least 10MB of monolingual data (AfroXLMR-61). 
To further extend to more languages with less than 10MB of data, we generate machine-translated data using NLLB for 34 African languages (including 18 in AfroXLMR-61). 
We refer to the resulting model after adaptation as AfroXLMR-76. We provide more details on the pre-training corpus in \autoref{appendix:maft_corpus}. 

\paragraph{Large Language Models} Lastly, we also report results by prompting two popular large language models: GPT-3.5-Turbo (gpt-3.5-turbo-0613) and GPT-4 (gpt-4-0613). Compared with smaller language models from MLM and MAFT, they feature strong instruction-following capabilities without task-specific fine-tuning.

\subsection{Training and evaluation scenarios}
\paragraph{Fully-supervised} In this setting, we trained on each language in \sib and evaluated on the same language. We did this evaluation for 205 languages and compared the performance of different text classification models. The MLP models were trained for 300 iterations, and we used either word \textit{ngram tokens} or \textit{XLM-R tokens}. 
For the multilingual PLM, we fine-tune each language training data for 20 epochs, with a maximum sequence length of 164, batch size of 16, and learning rate of 
$1e^{-5}$  on a single Nvidia A10 GPU. \textbf{Here, we assume access to labelled data in the target language.}

\paragraph{Cross-lingual transfer}
For this setting,  we \textbf{fine-tune} XLM-R on a language in Joshi's class 5 (we call it a “source” language), and \textbf{evaluate} on other languages. 
For this setting,  we \textbf{fine-tune} XLM-R on a language in Joshi's class 5 (we call it a “source” language), 
and \textbf{evaluate} on other languages. 
We trained in four languages with three different scripts i.e. English, French, Arabic and Chinese (Simplified). \textbf{Here, we assume access to labelled data in a few high-resource languages.}

\paragraph{Zero-shot prompt} We prompt GPT-3.5/4 for text classification for the 205 languages using an English template. We make use of a simple template from \citet{sanh2022multitask}:  \textit{`Is this a piece of news regarding \{\{``science, technology, travel, politics, sports, health, entertainment,  or geography''\}\}? \{\{INPUT\}\}'}. \textbf{Here, we assume no access to labelled data in any language}

\section{Results}

\begin{figure}[t]
    \centering
    \includegraphics[width=0.99\columnwidth]{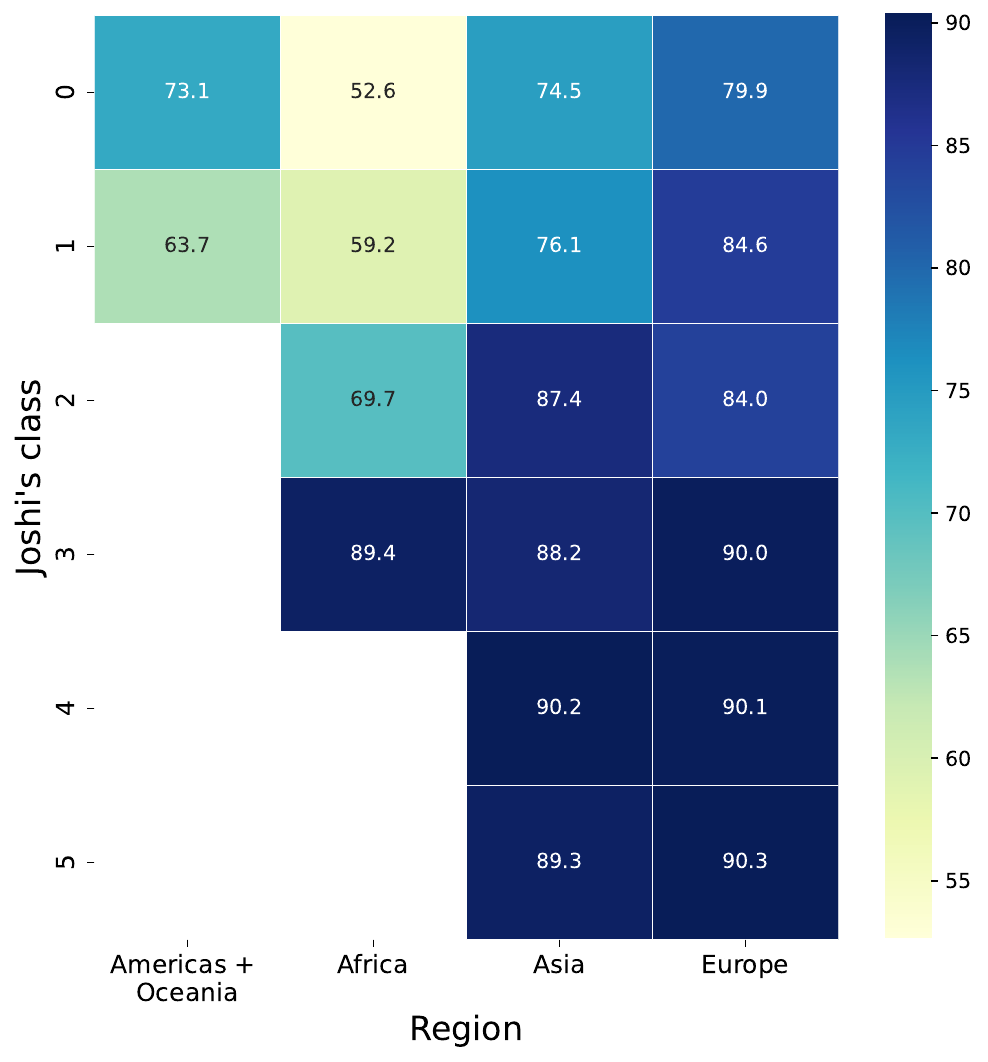}
    \vspace{-3mm}
    \caption{Heatmap of the performance by Region in each Joshi's class.} 
    \label{fig:heatmap_joshi_region}
\end{figure}

\begin{figure*}[t]
    \centering
    \includegraphics[width=2\columnwidth]{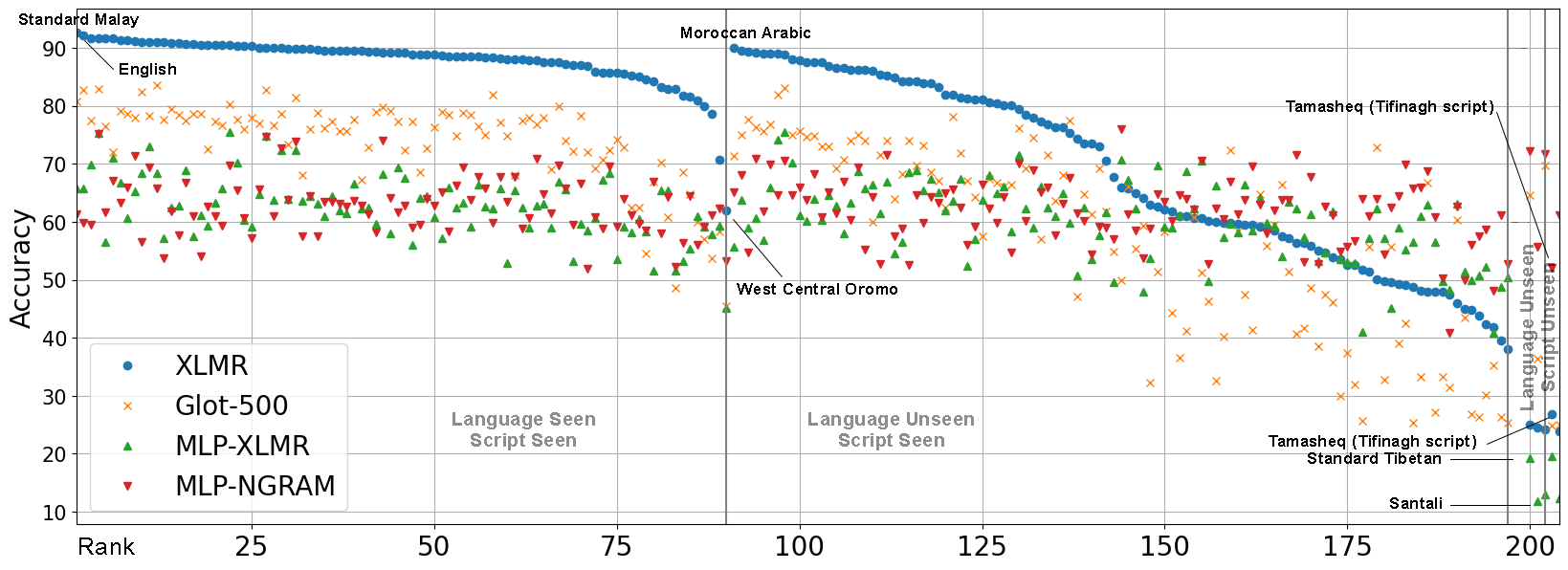}
    \vspace{-2mm}
    \caption{\textbf{Fully supervised Model Performance}. We group languages by whether they and their scripts are seen in the pre-training corpus of XLM-R. Languages are ordered by the XLM-R performance in every group.}
    \label{img:full_sup}
\end{figure*}

\subsection{Baseline results}

In order to demonstrate the effectiveness of our data set for multilingual evaluation, we benchmark the performance across various models and group the results by categorizations (\autoref{tab:baseline}). As XLM-R consistently outperforms Glot-500 across almost all language families, we use XLM-R as the baseline model in the cross-lingual transfer experiments.~\footnote{The full results are in \autoref{appendix:overall_result}}  Comparing English versus other languages, fine-tuning XLM-R on English achieved an accuracy of 92.1\%, indicating that \emph{the task itself is not difficult if given a properly pre-trained MLM and $\sim 700$ training samples}. However, when fine-tuning the same model in other languages, the performance drops vastly to an average accuracy of 75.9\%.   Similarly, in the cross-lingual transfer and zero-shot prompt scenarios, the performance further drops. 

\paragraph{Performances across language families}
The distribution of accuracy scores is imbalanced across language families. \emph{Atlantic-Congo, Nilotic, Mande, Aymaran and Quechuan languages have the lowest accuracy scores}. Even under the fully supervised scenario, the best-performed model reaches <65\% accuracy scores on these languages. There also tends to be a larger performance gap between fully-supervised and cross-lingual transfer scenarios, suggesting a poor semantic alignment~\cite{conneau2019cross} for these languages. Surprisingly, Tupian is the only additional language family that has >10\% drop from the fully supervised to cross-lingual transfer scenario. When moving further to the zero-shot prompt scenario, Basque shows the biggest performance drop (-36\%), next come the above-mentioned languages. Interestingly, despite this large decrease, Basque scores exceptionally high ($\approx$90\%) in the fully supervised and cross-lingual transfer scenarios.

\paragraph{Performances across Joshi's classes and geographical regions}
\autoref{fig:heatmap_joshi_region} visualizes the performance of XLM-R\footnote{We omit other models and only show XLM-R as ~\autoref{tab:baseline} has shown fine-tuning the XLM-R model performs the best} across different regions and Joshi's classes. We see a clear trend that languages with higher Joshi's classes perform better. Specifically, all languages with Joshi's class $\geq$3 have accuracy scores of $\approx$90\%. \emph{For languages in the same Joshi's class, African languages perform the worst, and European languages perform the best}. On Joshi's class 0, African languages are even at least 20\% worse than languages from other continents. Notably, there is no language with Joshi's class >3 in Africa and no American/Oceania
languages have Joshi's class >1. \emph{African and Oceania languages are also the only exceptions where MLP outperforms XLM-R}, implying a poorly learned representation of them. Future research should focus more on languages from these regions. \autoref{appendix:box_plot} provides the evaluation across the eight sub-regions instead of four in \autoref{fig:heatmap_joshi_region}. 

\paragraph{Performances across models}
In the fully supervised scenario, XLM-R performs the best on 16 out of the 22 language families. Among the remaining 6 language families, applying the simplest MLP classifier with n-gram input features outperforms more complex transformer-based MLMs (Glot-500 and XLM-R), suggesting they are not well adapted to these 6 language families. \emph{Glot-500, despite being pre-trained with many more languages, outperforms XLM-R only on Sino-Tibetan languages}. Even on Sino-Tibetan languages, it fails to out-perform the simplest MLP baseline. Cross-lingual transfer results are similar when using different source languages. On most language families, the results are comparable to fully supervised ones. Zero-shot prompting leads to a big drop due to the lack of supervised samples. The performance is good only for a few language families such as Indo-European, Uralic, Japonic and Koreanic.


\subsection{Factors affecting performance} 

In order to determine the critical factor in this multilingual classification task, we conducted in-depth case studies on the model architecture choices and language categorizations.

\paragraph{Effect of language coverage in pre-training} 
\autoref{img:full_sup} compares MLP, XLM-R and Glot-500 models based on language and script coverage in pre-training based on four groups: (1)  language seen, script seen in XLM-R (2) language unseen, script seen in XLM-R (3) script unseen in XLM-R, language seen in Glot-500 (4) script unseen by both models. The results in each group are sorted by their performance on fine-tuned XLM-R model.  Overall, \emph{XLM-R performs the best on all languages seen in its pre-training corpus without any exception}. 
Even for languages unseen in the pre-training corpus of XLM-R, it outperforms Glot-500 in most cases as long as the written scripts are seen. Glot-500 performs the best only for 3 out of all the 205 languages, implying their learned representations are far from sufficient. 
The reason could be that Glot-500 is pre-trained and evaluated on a religious corpus, which is quite different from the news domain in our task. In order to achieve a better generalization, we may have to mix text from various domains in the pre-training stage.

\paragraph{Effect of pre-training corpora size} 

\begin{figure}[h]
    \centering
    \includegraphics[width=0.95\columnwidth]{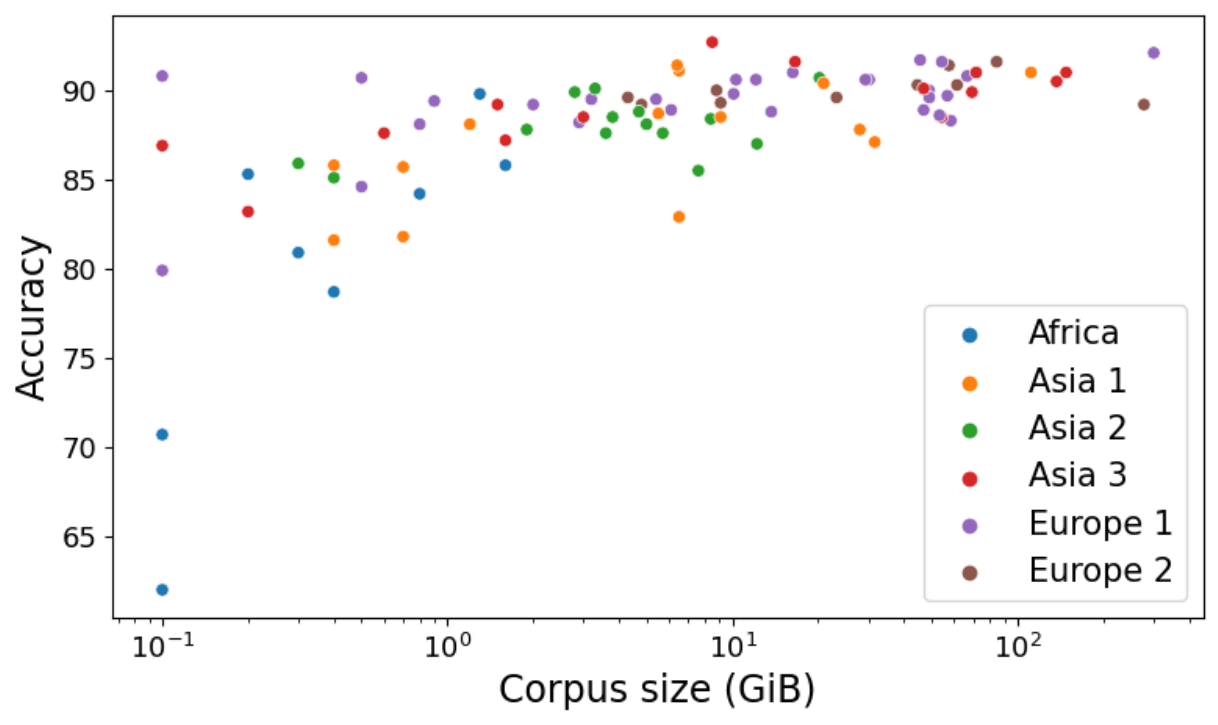}
    \vspace{-3mm}
    \caption{Accuracy of the XLM-R model vs Pre-Training corpus size in the fully supervised scenario. Bigger pre-training corpus in a target language generally improves the model performance.}
    \label{img:corpus_size}
\end{figure}


~\autoref{img:corpus_size} shows the change of accuracy scores with increasing corpus included in the pre-training stage of XLM-R, where the corpus size is logarithmically scaled for better visualization. We can see that \emph{with as little as 0.1GB pre-training corpus, the XLM-R model can already achieve $>$80\% accuracy for almost all languages}, which further verified that this task itself is not difficult. 
Though the accuracy generally grows with increasing corpus size, and the model performance starts to saturate with $>1$GB pre-training corpus. 

\begin{figure}[h]
    \centering
    \includegraphics[width=\columnwidth]{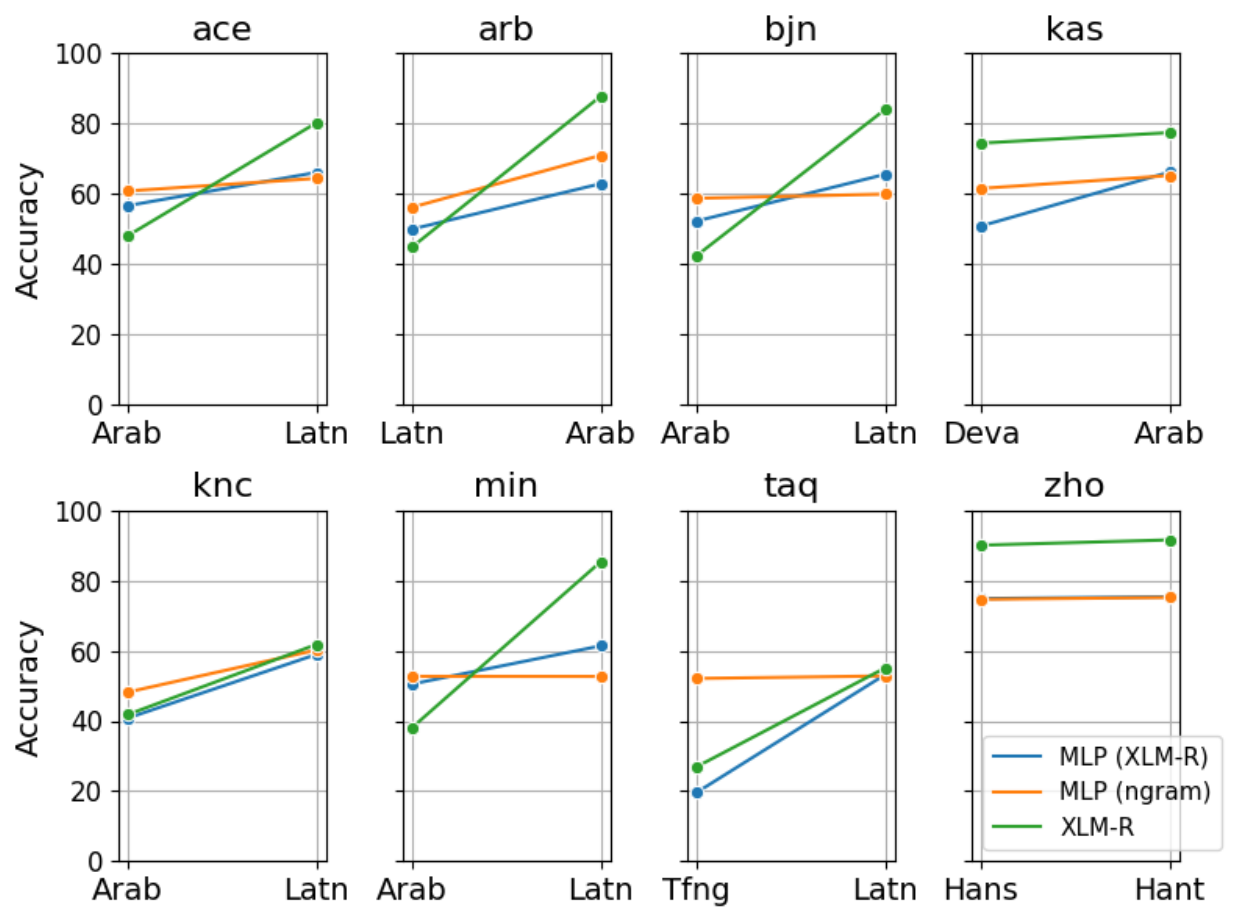}
    \vspace{-3mm}
    \caption{Script performance differences when one language has two different scripts. XLM-R and MLPs show the same trend. Using ngram features are more robust to script changes than using the XLM-R tokenizer.}
    \label{img:effect_script}
\end{figure}

\begin{figure*}[t]
    \centering
    \includegraphics[width=1.9\columnwidth]{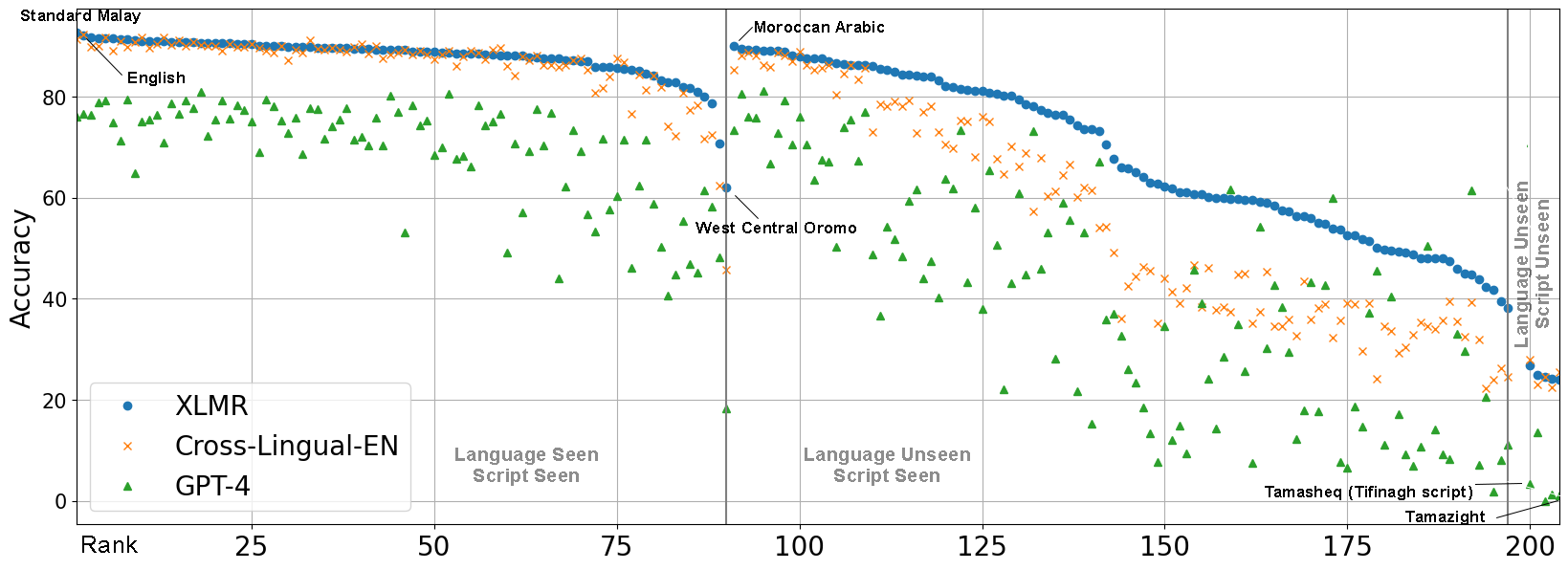}
    \vspace{-3mm}
    \caption{\textbf{Comparison of Various Scenarios}. We group languages by whether they and their scripts are seen in the pre-training corpus of XLM-R. Languages are ordered by the XLM-R fully-supervised performance in every group.}
    \label{img:scenario_modified}
\end{figure*}

\paragraph{Effect of script} 

To see how the choice of scripts affects the model performance, we choose eight languages that can be written in different scripts, and visualize the performance of XLM-R, MLP with n-gram features (MLP-ngram), and MLP with words from the XLM-R tokenizer (MLP-XLM-R) in ~\autoref{img:effect_script}. We can see that (1) The performance of MLP-XLMR usually correlates with that of XLM-R. This implies that under the XLM-R tokenizer, languages have their own preferred written scripts regardless of the effects from pre-training; (because this preference stays the same even with the simplest MLP classifier); 
(2) The slope of XLM-R is often steeper than that of MLP-XLMR, implying the preferred script for a language also has better pre-trained representations; (3) The slope of MLP-n-gram is often less steep. This implies that n-gram features are more robust across different scripts compared with word features obtained from the XLM-R tokenizer; (4) The preferred script is often the more commonly used one for every language.~\footnote{We define ``preferred written scripts'' as the writing system or script that individuals or communities predominantly choose or favor when expressing written language. } 

\subsection{Comparison of different scenarios} 
\paragraph{Fine-tune vs. Prompted} 
Out of all the 205 languages, GPT-4 outperforms GPT-3.5-turbo in 157 languages. Only on Buginese, Kabiyè, Mizo, Nuer and Ayacucho Quechua, GPT-3.5-Turbo outperforms GPT-4 for $>10\%$. However, zero-shot prompting consistently underperforms fine-tuned methods. 
It is hard to include extensive descriptions of the classification criteria in the prompt. Adding more examples to the prompt might improve the performance. 

\paragraph{Cross-Lingual transfer vs Fully supervised}
Here, we compare the performance between cross-lingual transfer and fully-supervised methods. We observe that all languages that are included in the pre-training corpus of XLM-R, the cross-lingual transfer performs similarly to fully supervised methods. The best source language for cross-lingual transfer is, surprisingly, French, rather than English, which has the largest amount of pre-training corpus, though the difference among various source languages is tiny. 
This suggests \emph{languages included in the XLM-R pre-training corpus are pretty well aligned with all the four chosen high-resource languages}. 
The advantage of fully supervised methods over cross-lingual transfer becomes prominent mainly when the target language is not included in the pre-training corpus of XLM-R and its script is included. In this case, fully supervised methods can improve the performance by fine-tuning the model on the target languages, but cross-lingual transfer fails to capture the alignment with high-resource languages. 
\autoref{img:scenario_modified} provides comparison between cross-lingual transfer and fully-supervised methods, including GPT-4 evaluation.


\subsection{Region-specific pre-training}

\paragraph{Evaluation of region-specific PLMs} While our evaluation is primarily focused on multilingual PLMs trained on 100 languages or more, models pre-trained on a group of linguistically or geographically related languages often lead to better performance as observed for Indian languages (\autoref{tab:indic_bert}) and African languages (\autoref{tab:african_bert}). IndicBERTv2 and MurilBERT achieved better overall performance over XLM-R (550M parameters) despite their smaller capacity (236M-278M parameters), especially for Indian languages they both support, and better for languages not covered by XLM-R. Similarly for African languages, AfroXLMR---an adaptation of XLM-R through multilingual adaptive fine-tuning (MAFT)~\citep{alabi-etal-2022-adapting} to 17 African languages gave 
roughly $+9$ improvement in performance. AfriBERTa on the other hand slightly gave worse result than XLM-R despite seeing the same number of African languages during pre-training (although not the exact languages) because it was pre-trained on less amount of data (1GB). Despite the improvement of AfroXLMR, it performs terribly for Nilotic, Mande and many Atlantic Congo families which shows that including more African languages in pre-training could improve performance. 

\paragraph{Performance of applying MAFT to more African languages} We evaluated on two MAFT models described in (\S\ref{sec:text_class_models}). Our evaluation of AfroXLMR-76 shows that MAFT with synthetic data was effective in improving the accuracy over AfroXLMR in many languages in Africa, especially for Nilotic ($+7.9$), Mande ($+4.5$) and Atlantic-Congo ($+7.4$) languages, similar to the findings of \citet{urbizu-etal-2023-enough}. 
The performance improvement for AfroXLMR-61 was smaller on average ($+3.4$). There are few cases where it leads to a slight drop in performance on more-resource languages due to curse-of-multilinguality~\citep{conneau-etal-2020-unsupervised}. The newly developed PLMs are available on HuggingFace.\footnote{\url{https://huggingface.co/Davlan}} 
Despite the improvement in performance, African languages whose script were not covered by the XLM-R tokenizer (like N'ko and Tamazight) did not improve. 
To address this issue, 
we provide an extension of AfroXLMR-76 with vocabulary augumentation in \autoref{appendix:african_result}. 

 


\begin{table}[t] 
\centering
\resizebox{\columnwidth}{!}{%
\begin{tabular}{lrrrr|r}
\toprule
 & \multicolumn{4}{c}{\textbf{Language Family}} \\
\multirow{3}{*}{\textbf{Models}} & \textbf{Indo-E} & \textbf{Dravidian} & \textbf{Austro-Asia} & \textbf{Sino-Tib} & \textbf{All} \\ 
& \textbf{(18)} & \textbf{(4)} & \textbf{(1)} & \textbf{(1)} & \textbf{(24)} \\
\midrule
XLM-R  & 86.5 & 87.9 & 24.6 & 48.7 &  82.6 \\ 
IndicBERTv2  & 85.4 & 88.3  & \textbf{65.5} & 43.2  & 83.3 \\  
MurilBERT  & \textbf{87.5} & \textbf{89.9} & 23.5 & \textbf{66.3} & \textbf{84.4} \\ 
\bottomrule
\end{tabular}
}
\vspace{-2mm}
\caption{\textbf{Indic-centric Evaluation on \textsc{SIB-200}}. }
\label{tab:indic_bert}

\end{table}

\begin{table}[t] 
\centering
\resizebox{\columnwidth}{!}{%
\begin{tabular}{lrrrrrr|r}
\toprule
 & \multicolumn{6}{c}{\textbf{Language Family}} \\
\multirow{3}{*}{\textbf{Models}} & \textbf{A.Congo} & \textbf{Afro A.} & \textbf{Nilo.} & \textbf{Mande} & \textbf{Aust.} & \textbf{Indo-E} & \textbf{All} \\ 
& \textbf{(34)} & \textbf{(12)} & \textbf{(5)} & \textbf{(3)} & \textbf{(1)} & \textbf{(1)} & \textbf{(56)} \\
\midrule
MLP  & 61.3 & 59.6 & 58.6 & \textbf{61.7} & 61.1 & 57.6 & 60.6 \\ 
AfriBERTa  & 58.8 & 50.9 & 54.2 & 40.4 & 50.5 & 53.7 & 55.4 \\ 
XLM-R  & 57.9 & 65.4 & 53.7 & 40.2 & 85.3 & 89.8 & 59.2  \\ 
Serengeti & 65.1 & 62.4 & 53.5 & 38.7 & 80.7 & 86.9 & 62.7 \\
AfroXLMR  & 70.8 & 69.2 & 55.7 & 45.6 & \textbf{88.4} & 90.4 & 68.4 \\ 
\midrule
AfroXLMR-61  & 74.8 & 68.3 & 57.2 & 44.8 & 88.2 & 89.1 & 70.7 \\ 
AfroXLMR-76  & \textbf{78.2} & \textbf{69.9} & \textbf{63.6} & 50.1 & 88.1 &  \textbf{91.1} & \textbf{74.1} \\ 
\bottomrule
\end{tabular}
}
\vspace{-2mm}
\caption{\textbf{African-centric Evaluation on \textsc{SIB-200}}. }
\label{tab:african_bert}

\end{table}

\section{Related Work}


There have been several efforts to curate multilingual evaluation datasets, including various downstream tasks such as part-of-speech tagging~\citep{nivre-etal-2016-universal, nivre-etal-2020-universal, dione-etal-2023-masakhapos}
, named entity recognition%
~\citep{pan-etal-2017-cross,adelani-etal-2022-masakhaner,mhaske-etal-2023-naamapadam}, entity linking~\cite{botha-etal-2020-entity}, natural language inference~\citep{conneau-etal-2018-xnli}, text classification~\cite{fitzgerald-etal-2023-massive,Ma2023Taxi1500AM}, machine translation%
~\citep{ goyal-etal-2022-flores, team2022NoLL,adelani-etal-2022-thousand}, and question answering
~\cite{lewis-etal-2020-mlqa,shen2023-xpqa,doddapaneni-etal-2023-towards,bandarkar2023belebele}. All these initiatives have played a pivotal role in advancing the field of cross-lingual and multilingual NLP. Our work, which focuses on the creation of an extensive multilingual text classification dataset covering 200 languages, builds upon a line of related works that have contributed to the expansion of the NLP community. 

Specific to text classification, a few multilingual datasets are IndicNLP BBC news~\citep{kunchukuttan2020indicnlpcorpus}, KINNEWS \& KIRNEWS~\citep{niyongabo-etal-2020-kinnews}, ANTC~\cite{alabi-etal-2022-adapting}, MasakhaNEWS~\citep{adelani2023masakhanews},  and Taxi1500~\citep{Ma2023Taxi1500AM}. To the best of our knowledge, Taxi1500 is the most recent and largest of them all covering 1500 languages. However, this dataset is focused on the religious domain as the data comes from the Bible. Our work addresses a gap in multilingual text classification datasets by curating \sib that covers a broader range of topics and domains.

\section{Conclusion}
In this paper, we created SIB-200---a large scale open-sourced benchmark dataset for topic classification in 200 languages and dialects to address the lack of evaluation datasets for natural language understanding especially for low-resource languages. We performed extensive evaluation across full-supervised setting, cross-lingual transfer setting and prompting of LLMs settings. Furthermore, we grouped the 200 languages in different categories based on language families, geographical regions, Joshi's class and coverage in multilingual pre-trained language models to provide insights into which group of languages have poor performance on this simple and inclusive benchmark. 



%

\section{Limitations}
\paragraph{Data size} One of the limitations of our work is the size of the benchmark data which is 1,004. Having more instances would be better.  However, we believe this is an important contribution for many languages that often do not have dataset (e.g. news articles or Wikipedia articles) that can be used for topic classification annotation. 

\paragraph{Translationese effect}
One of the main limitation of our work is that the labelled dataset created for other non-English languages are based on human translation and may suffer from translationese effect including a slight drop in performance.

\paragraph{Few PLMs evaluated}
Another limitation is the choice of multilingual pre-trained language models, we note that XLM-R may not be the best multilingual encoder out there, there are other publicly available ones like InfoXLM~\citep{chi-etal-2021-infoxlm}, mDeBERTa~\citep{he2023debertav} and others, however due to the scale of the experiments, we limited our evaluation to three multilingual models (XLM-R-base, XLM-R, and Glot-500). We believe our result may still be consistent with newer PLMs since they often cover similar set of languages as XLM-R. 

\section{Acknowledgement}
David Adelani acknowledges the support of DeepMind Academic Fellowship programme. Jesujoba Alabi was partially funded by the BMBF project SLIK under the Federal Ministry of Education and Research grant 01IS22015C. This work was supported in part by Oracle Cloud credits and related resources provided by Oracle. We thank Google for providing GCP credits to train the AfroXLMR-61 and AfroXLMR-76 models. Finally, we are grateful to OpenAI for providing API credits through their Researcher Access API programme to Masakhane for the evaluation of GPT-3.5 and GPT-4 large language models.

\bibliography{anthology,custom}
\bibliographystyle{acl_natbib}

\appendix

\section{Languages and their categorizations}
\label{appendix:plms_covered}

\autoref{tab:languages} and \autoref{tab:lang_family_plm} shows the grouping of languages in the \sib dataset. We categorized them based on the following characteristics:  
(1) geographical regions, (2) language family, (3) coverage in multilingual PLMs, and (4) Joshi's classification~\citep{joshi-etal-2020-state}---a  categorization based on their labelled/unlabelled resources on the web---making it easy to analyze results.

\paragraph{Categorization by geographical regions} \autoref{tab:languages} shows the grouping of languages into regions according to the United Nations Geoscheme\footnote{\url{https://en.wikipedia.org/wiki/United_Nations_geoscheme}}. The regions are: Africa, Americas, Asia 1 or Western \& Central Asia, Asia 2 or Southern Asia, Asia 3 or South-Eastern \& Eastern Asia, Europe 1 or Northern/Western/Southern Europe, Europe 2 or Eastern Europe, and Oceania. Asia, Europe, and Africa regions have the largest number of languages with 82, 57, and 56 languages respectively. The Oceania and the Americas regions have the lowest number of languages with four and five respectively. 

\paragraph{Categorization by language family} \sib languages are grouped into 21 language families as shown in \autoref{tab:lang_family_plm}, the largest groups are: Indo-European (79 languages), Atlantic-Congo (34 languages), Afro-Asiatic (21 languages), Austronesian (21 languages) and Turkic (11 languages).

\paragraph{Categorization by Joshi's classification} \autoref{tab:languages} also shows the number of languages in each Joshi's class---a measure of the unlabelled or labelled resources available for each language on the web~\citep{joshi-etal-2020-state}. 128 languages can be categorized as low-resource since they fall between class ``0'' and ``2'', 30 languages are mid-resource in class ``3'', and the others are high-resource (only 39 languages). This also corresponds to the NLLB classification for machine translation resources available on the web, but with only two categories---150 low-resource languages and 54 high-resource languages. 

\paragraph{Categorization by availability in PLM} Lastly, we grouped languages and language families by their inclusion in the training of multilingual PLMs. XLM-R~\citep{conneau-etal-2020-unsupervised} covered 90 out of the 205 languages in our dataset while GLOT-500~\citep{imanigooghari-etal-2023-glot500} covered 177. This is a good indication of performance in general since languages that are included during pre-training often have better performance~\citep{ponti-etal-2020-xcopa,pfeiffer-etal-2020-mad,adelani-etal-2022-masakhaner}. 
Finally, we show the number of languages covered by region-specific PLMs such as AfriBERTa~\citep{ogueji-etal-2021-small}, AfroXLMR~\citep{alabi-etal-2022-adapting}, MuRIL~\citep{Khanuja2021MuRILMR}, and IndicBERTv2~\citep{doddapaneni-etal-2023-towards}. The grouping is provided in \autoref{tab:lang_family_plm}.


\section{Pre-training corpus for MAFT}
\label{appendix:maft_corpus}
We explore how to improve over regional PLMs using MAFT---adaptation of an existing multilingual PLM to multiple or new set of languages simultaneously, this was effective for adapting XLM-R to 20 languages spoken in Africa~\citep{alabi-etal-2022-adapting}. To extend to more languages, we apply MAFT to 61 African languages with at least 10MB of monolingual data (AfroXLMR-61). The data was obtained from the concatenation of different web sources like AfroXLMR training corpus, \textbf{MT560}~\citep{gowda-etal-2021-many} (mostly religious articles), \textbf{Flores-200 (multi-domain)}, and \textbf{Wikipedia}. In total, this results in \texttt{17GB} of data. 

To further extend to more languages with less than 10MB of data, \textit{we generate machine-translated data using NLLB for 34 African languages} (including 18 in AfroXLMR-61). The selected 34 languages are the ones with less than 10MB or only have MT560 (religious domain). We make use of the \textbf{English news commentary dataset}\footnote{we used version 16 of the data released for WMT.}~\citep{kocmi-etal-2022-findings} with over 600,000 sentences to translate to these 34 languages. 
We refer to the resulting model after adaptation as AfroXLMR-76 which has been pre-trained on \texttt{21GB} of data.

\begin{table}[t] 
\centering
\resizebox{\columnwidth}{!}{%
\begin{tabular}{lrrrrrrr}
\toprule
\textbf{Language} & \textbf{\#Lang} & &  \textbf{Glot} & \textbf{Afri} & \textbf{Afro} & \textbf{Indic}  \\ 
\textbf{Family} & \textbf{SIB} & \textbf{XLM-R} & \textbf{500} & \textbf{BERTa} & \textbf{XLM-R} & \textbf{BERTv2}  \\ 
\midrule
Indo-European & 79 & 50 & 72 & 1 & 4  & 14 \\ 
Atlantic-Congo & 34 & 2 & 32 & 6 & 10   & -\\ 
Afro-Asiatic & 21& 6 & 14 & 4 & 5 &  - \\ 
Austronesian & 21 & 5 & 17 & - & 1 &  -\\ 
Turkic & 11& 7 & 11 & - & - &  -\\ 
Sino-Tibetan &9  & 3 & 7 & - & - &   -\\ 
Nilotic  & 5 & - & 1 & - & -&  -\\ 
Dravidian & 4 & 4 & 4 & - & - & 4 \\
Tai-Kadai & 3 & 2 & 2 & - & -  & - \\
Uralic & 3 &  3 & 3 & - & - & -\\
Austroasiatic & 3 & 2 & 3 & - & -& 1\\
Mande & 3 & - & 2 & - & - &  -\\
Japonic & 1 & 1 & 1 & - & -&  -\\
Koreanic & 1 & 1 & 1 & - & - &  - \\
Mongolic-Khitan & 1 & 1 & 1 & - & - & -\\
Constructed & 1 & 1 & 1 & - & - & -\\
Quechuan & 1 & - & 1 & - & - & -\\
Basque & 1 & 1 & 1 & - & - & - \\
Aymaran & 1 & - & 1 & - & -  & - \\
Tupian & 1 & - & 1 & - & - & - \\
Kartvelian & 1 & 1 & 1 & - & - & -\\
\midrule
Total & 205 & 90 & 177 & 11 & 20 & 19 \\

\bottomrule
\end{tabular}
}
\caption{\textbf{Languages covered in multilingual pre-trained language model} and their language families. We excluded MuRIL because it was trained in similar languages as IndicBERTv2 except for Santali in the Austroasiatic family.}
\label{tab:lang_family_plm}

\end{table}

\section{SIB-200 English dataset performance using 7 or 14 labels}
\label{appendix:eng_result}
By fine-tuning XLM-R SIB-200 with 14 labels, we achieved accuracy score of 82.3\% while for the 7 labels, we reached the performance of 92.5\%.

\section{Overall result}
\label{appendix:overall_result}
\autoref{tab:all_language_results} shows the overall results for all languages. 

\section{Results by categorization of regions}
\label{appendix:box_plot}
\begin{figure*}[h]
    \centering
    \includegraphics[width=2\columnwidth]{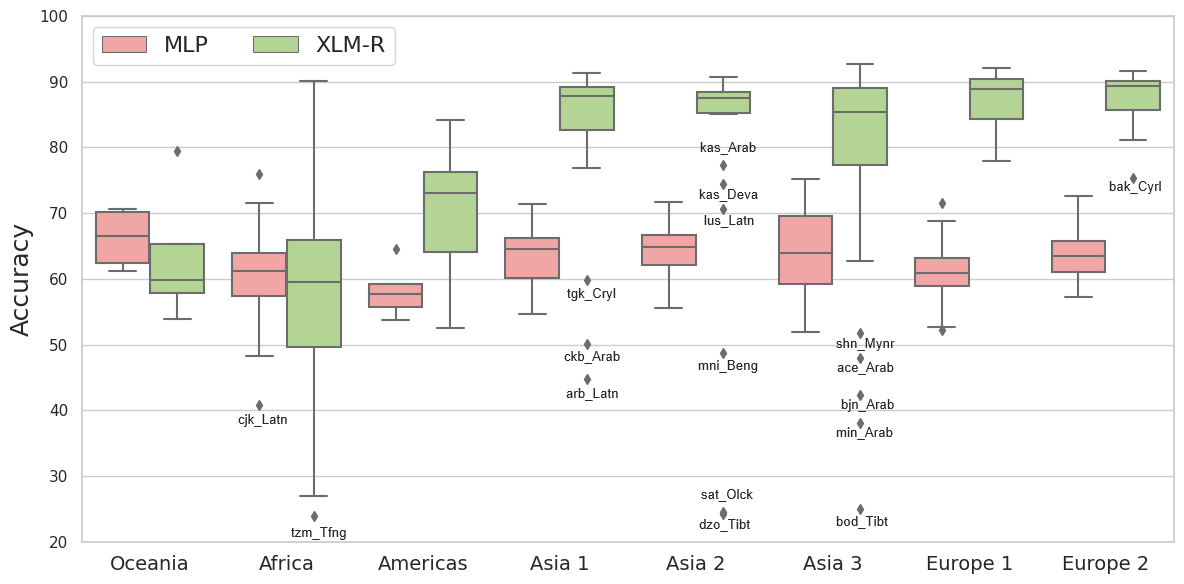}
    \caption{Box plot comparison of MLP and XLM-R model across regions}
    \label{fig:box_plot}
\end{figure*}

\autoref{fig:box_plot} shows the baseline results of each region represented in a box plot. 

\section{African languages result}
\label{appendix:african_result}

\paragraph{Vocabulary augmentation}to address the non-coverage of some African scripts like \texttt{Nkoo} and \texttt{Tfng}, we perform vocabulary augumentation of the original XLM-R tokenizer. We follow these steps: (1) we train a tokenizer on a combined multilingual texts for N'ko, Tamasheq (Tifinagh) and Tamazight languages using sentencepiece, and vocabulary size of 30K. (2) We added the top-20K new vocabulary tokens to the XLM-R vocabulary. (3) We performed MAFT on XLMR. The resulting model is called AfroXLMR-76-script. As an additional experiment, we repeated the vocabulary augmentation and MAFT approach for XLM-R-base model, resulting into AfroXLMR-base-76-script. 

\paragraph{Results categorized by script} Our results in \autoref{tab:african_bert_script} shows that vocabulary augumentation was effective for the languages that use \texttt{Nkoo} and \texttt{Tfng} with over $+18$ points improvement but the performance is still lower than using MLP for these languages.  This analysis shows the importance of building PLMs with diverse scripts during the pre-training phase. 

\begin{table}[t] 
\centering
\resizebox{\columnwidth}{!}{%
\begin{tabular}{lrrrrr|r}
\toprule
 & \multicolumn{6}{c}{\textbf{Script}} \\
\multirow{3}{*}{\textbf{Models}} & \textbf{Latn} & \textbf{Arab} & \textbf{Ethi} & \textbf{Tfng} & \textbf{Nkoo} & \textbf{All} \\ 
& \textbf{(47)} & \textbf{(4)} & \textbf{(2)} & \textbf{(2)} & \textbf{(1)} & \textbf{(56)} \\
\midrule
MLP  & 60.6 & 60.5 & 62.0 & 56.6 & 70.3 & 60.6 \\ 
XLM-R & 59.2 & 76.9 & 76.0 & 25.4 & 23.2 & 59.2  \\ 
\midrule
AfriBERTa  & 58.8 & 28.5 & 75.3 & 26.5 & 22.2 & 55.4 \\ 
Serengeti & 64.2 & 68.0 & 75.7 & 24.8 & 22.0 & 62.7 \\
AfroXLMR-base  & 63.6 & 74.0 & 76.9 & 27.7 & 21.2 & 62.7 \\ 
AfroXLMR  & 69.6 & 77.4 & \textbf{85.1} & 26.9 & 22.7 & 68.4 \\ 
\midrule
AfroXLMR-61  & 72.5 & 77.3 & 84.1 & 25.5 & 22.5 & 70.7 \\ 
AfroXLMR-76  & \textbf{76.5} & \textbf{77.5} & 83.5 & 26.2 & 22.0 & 74.1 \\ 
AfroXLMR-76-script  & 75.6 & \textbf{77.5} & 82.2 & \textbf{45.8} & \textbf{40.5} &  \textbf{74.3} \\ 
AfroXLMR-base-76-script  & 68.4 & 74.2 & 72.1 & 41.0 & 39.1 &  67.4 \\ 

\bottomrule
\end{tabular}
}
\vspace{-2mm}
\caption{\textbf{African-centric Evaluation based on script on \textsc{SIB-200}}. The number of languages per script are in brackets. }
\label{tab:african_bert_script}

\end{table}

\paragraph{Overall African languages evaluation}
\autoref{tab:african_result_table} shows the overall results for the African languages.

\onecolumn

\begin{scriptsize}\setlength\tabcolsep{3pt}
    \captionsetup{width=15cm}

    \begin{longtable}{p{25mm}lllll|rrr|rrr|rr}

        & & & & & & \multicolumn{3}{c|}{\textbf{Fully Supervised}} &\multicolumn{3}{c|}{\textbf{Cross-lingual}} & \multicolumn{2}{c}{\textbf{Prompting}}\\
        & \textbf{Language} & \textbf{Joshi's} &  \textbf{in} & \textbf{Language} &  &  & \textbf{XLM-R} & & \multicolumn{3}{c|}{\textbf{XLM-R-based}}& &  \\ 
        \textbf{Language name} & \textbf{code} & \textbf{class} & \textbf{XLM-R?} & \textbf{family} & \textbf{region} & \textbf{MLP} & \textbf{base} & \textbf{XLM-R} & \textbf{eng} & \textbf{ara} & \textbf{zho} & \textbf{GPT-3.5} & \textbf{GPT-4} \\ 
        \midrule
        Acehnese (Arabic) & ace\_Arab & 1 & ~ & Austronesian & Asia 3 & \textbf{60.77} & 35.7 & 48 & 34 & 32.5 & 33.2 & 9.1 & 14.2 \\ 
        Acehnese (Latin) & ace\_Latn & 1 & ~ & Austronesian & Asia 3 & 64.35 & 74 & \textbf{80.2} & 64.7 & 67.4 & 67.4 & 31.4 & 22.1 \\ 
        Mesopotamian Arabic & acm\_Arab & ~ & ~ & Afro-Asiatic & Asia 1 & 68.16 & 86.9 & 89.5 & 88.1 & 88.8 & \textbf{89.7} & 62.6 & 80.6 \\ 
        Ta’izzi-Adeni Arabic & acq\_Arab & ~ & ~ & Afro-Asiatic & Asia 1 & 65.97 & 87.5 & 87.9 & 88.8 & \textbf{89.4} & 89.1 & 63 & 76 \\ 
        Tunisian Arabic & aeb\_Arab & ~ & ~ & Afro-Asiatic & Africa & 66.85 & 83.9 & \textbf{86.5} & 84.6 & 85.7 & 85.3 & 54.2 & 73.9 \\ 
        Afrikaans & afr\_Latn & 3 & \cmark & Indo-European & Africa & 57.55 & 87.8 & \textbf{89.8} & 88.7 & 88.6 & \textbf{89.8} & 63.6 & 68.6 \\ 
        South Levantine Arabic & ajp\_Arab & ~ & ~ & Afro-Asiatic & Asia 1 & 60.28 & 83.1 & 86.3 & 86.2 & \textbf{87.1} & \textbf{87.1} & 63 & 75.4 \\ 
        Akan & aka\_Latn & 1 & ~ & Atlantic-Congo & Africa & \textbf{61.33} & 53.7 & 59.7 & 44.8 & 42.6 & 45 & 31.1 & 35 \\ 
        Tosk Albanian & als\_Latn & 1 & \cmark & Indo-European & Europe 1 & 63.6 & 87.7 & 89.5 & 89.8 & \textbf{90.5} & \textbf{90.5} & 66.5 & 71.4 \\ 
        Amharic & amh\_Ethi & 2 & \cmark & Afro-Asiatic & Africa & 66.87 & 80.1 & 84.2 & 84.1 & \textbf{85.7} & 83.1 & 23.4 & 58.8 \\ 
        North Levantine Arabic & apc\_Arab & ~ & ~ & Afro-Asiatic & Asia 1 & 54.66 & 85.1 & 89.3 & 88.6 & \textbf{89.4} & 89 & 59.9 & 76 \\ 
        MSA (Arabic) & arb\_Arab & 5 & \cmark & Afro-Asiatic & Asia 1 & 70.93 & 88.2 & 87.8 & 88.1 & \textbf{89.1} & 88.9 & 68.4 & 77.6 \\ 
        MSA (Romanized) & arb\_Latn & ~ & ~ & Afro-Asiatic & Asia 1 & 56.12 & 36.3 & 44.8 & 39.3 & 39.6 & 39.9 & 44.7 & \textbf{61.5} \\ 
        Najdi Arabic & ars\_Arab & ~ & ~ & Afro-Asiatic & Asia 1 & 70.87 & 88.9 & \textbf{89.2} & 88.2 & 89.1 & 89.1 & 67 & 75.9 \\ 
        Moroccan Arabic & ary\_Arab & ~ & ~ & Afro-Asiatic & Africa & 65.09 & 83 & \textbf{90.1} & 85.3 & 85.8 & 85.6 & 47.3 & 73.3 \\ 
        Egyptian Arabic & arz\_Arab & 3 & ~ & Afro-Asiatic & Africa & 61.82 & 84.8 & \textbf{89.1} & 86.2 & 88 & 87.9 & 58 & 81.1 \\ 
        Assamese & asm\_Beng & 1 & \cmark & Indo-European & Asia 2 & 63.4 & 75.5 & \textbf{88.1} & 86.1 & 85.2 & 85.6 & 51.8 & 49.2 \\ 
        Asturian & ast\_Latn & 1 & ~ & Indo-European & Europe 1 & 63.71 & 85.8 & \textbf{87.5} & 86.3 & 86.1 & 86.1 & 58.5 & 70.6 \\ 
        Awadhi & awa\_Deva & 0 & ~ & Indo-European & Asia 2 & 64.68 & 85 & 88.1 & 87 & \textbf{88.8} & 88.1 & 59.7 & 70.6 \\ 
        Central Aymara & ayr\_Latn & 1 & ~ & Aymaran & Americas & \textbf{55.68} & 42.5 & 52.5 & 39.1 & 41.3 & 38.5 & 15.9 & 6.6 \\ 
        South Azerbaijani & azb\_Arab & 1 & \cmark & Turkic & Asia 1 & 64.27 & 79 & \textbf{82.9} & 74.1 & 73.6 & 74.1 & 32.9 & 40.6 \\ 
        North Azerbaijani & azj\_Latn & 1 & \cmark & Turkic & Asia 1 & 71.4 & 85 & \textbf{91.1} & 90.8 & 90.3 & \textbf{91.1} & 57.2 & 64.8 \\ 
        Bashkir & bak\_Cyrl & 1 & ~ & Turkic & Europe 2 & 67.51 & 72.2 & \textbf{75.4} & 66.5 & 70.2 & 68.6 & 45 & 55.6 \\ 
        Bambara & bam\_Latn & 1 & ~ & Mande & Africa & \textbf{64.43} & 42.1 & 49.3 & 29.2 & 29 & 29.5 & 23.6 & 17.2 \\ 
        Balinese & ban\_Latn & 0 & ~ & Austronesian & Asia 3 & 64.3 & 79 & \textbf{83.9} & 78 & 79.7 & 76.9 & 51.7 & 47.4 \\ 
        Belarusian & bel\_Cyrl & 3 & \cmark & Indo-European & Europe 2 & 63.48 & 86.4 & \textbf{89.6} & \textbf{89.6} & 89.3 & 89.3 & 54.3 & 74.2 \\ 
        Bemba & bem\_Latn & 0 & ~ & Atlantic-Congo & Africa & \textbf{63.76} & 52.9 & 59.5 & 44.9 & 45.9 & 45.3 & 21.8 & 25.6 \\ 
        Bengali & ben\_Beng & 3 & \cmark & Indo-European & Asia 2 & 65.76 & 83 & \textbf{88.4} & 87.4 & 87.5 & 85.6 & 59.6 & 74.4 \\ 
        Bhojpuri & bho\_Deva & 1 & ~ & Indo-European & Asia 2 & 69.37 & 82 & \textbf{86.3} & 83.3 & 83.2 & 84.1 & 49.4 & 67.4 \\ 
        Banjar (Arabic script) & bjn\_Arab & 1 & ~ & Austronesian & Asia 3 & \textbf{58.68} & 36.6 & 42.3 & 22.2 & 21.8 & 22.5 & 12.1 & 20.5 \\ 
        Banjar (Latin script) & bjn\_Latn & 1 & ~ & Austronesian & Asia 3 & 59.87 & 79.8 & \textbf{84} & 77 & 78.5 & 76.8 & 34.2 & 44.1 \\ 
        Standard Tibetan & bod\_Tibt & 1 & ~ & Sino-Tibetan & Asia 3 & \textbf{72.14} & 24.8 & 25 & 23 & 20.9 & 20.7 & 5.8 & 13.6 \\ 
        Bosnian & bos\_Latn & 3 & \cmark & Indo-European & Europe 1 & 57.68 & 87.8 & 90.8 & 91.1 & 90.4 & \textbf{92.3} & 65.7 & 76.6 \\ 
        Buginese & bug\_Latn & 1 & ~ & Austronesian & Asia 3 & 54.38 & 72.6 & \textbf{73.5} & 61.5 & 63.1 & 64.1 & 26.9 & 15.2 \\ 
        Bulgarian & bul\_Cyrl & 3 & \cmark & Indo-European & Europe 2 & 66.01 & 88.4 & \textbf{91.4} & 89.9 & 89.3 & 90.1 & 66.5 & 79.4 \\ 
        Catalan & cat\_Latn & 4 & \cmark & Indo-European & Europe 1 & 64.49 & 88.6 & 89.8 & 91.1 & 91.4 & \textbf{91.8} & 55.4 & 77.7 \\ 
        Cebuano & ceb\_Latn & 3 & ~ & Austronesian & Asia 3 & 62.5 & 77.9 & \textbf{81.5} & 75.3 & 77.2 & 76.6 & 62.3 & 73.4 \\ 
        Czech & ces\_Latn & 4 & \cmark & Indo-European & Europe 1 & 53.67 & 88.7 & 91 & 91.8 & 91 & \textbf{92.3} & 62.4 & 70.9 \\ 
        Chokwe & cjk\_Latn & ~ & ~ & Atlantic-Congo & Africa & 40.83 & 43.8 & \textbf{47.5} & 39.5 & 40.8 & 39.9 & 14.3 & 8.2 \\ 
        Central Kurdish & ckb\_Arab & 0 & ~ & Indo-European & Asia 1 & \textbf{63.99} & 37.7 & 50.1 & 24.1 & 25.7 & 24.9 & 45.6 & 45.6 \\ 
        Crimean Tatar & crh\_Latn & 1 & ~ & Turkic & Europe 2 & 61.48 & 80.9 & \textbf{86.6} & 80.3 & 82.2 & 81.3 & 34.5 & 50.2 \\ 
        Welsh & cym\_Latn & 1 & \cmark & Indo-European & Europe 1 & 67.76 & 81.3 & \textbf{88.1} & 84.2 & 84.5 & 83.8 & 59.4 & 70.7 \\ 
        Danish & dan\_Latn & 3 & \cmark & Indo-European & Europe 1 & 59.42 & 88 & \textbf{91.7} & 89.8 & 90 & 91 & 69.6 & 76.3 \\ 
        German & deu\_Latn & 5 & \cmark & Indo-European & Europe 1 & 61.81 & 88.4 & 90.8 & 90.2 & 91 & \textbf{91.1} & 70.7 & 78.6 \\ 
        Southwestern Dinka & dik\_Latn & 1 & ~ & Nilotic & Africa & \textbf{64.58} & 51.1 & 61 & 39.2 & 40 & 38.4 & 24 & 14.9 \\ 
        Dyula & dyu\_Latn & 0 & ~ & Mande & Africa & \textbf{50.34} & 43.3 & 48 & 35.8 & 35.1 & 35 & 12.3 & 9.3 \\ 
        Dzongkha & dzo\_Tibt & 1 & ~ & Sino-Tibetan & Asia 2 & \textbf{71.66} & 26 & 24.2 & 22.4 & 20 & 20 & 0 & 1.2 \\ 
        Greek & ell\_Grek & 3 & \cmark & Indo-European & Europe 1 & 59.04 & 85.5 & 88.9 & 88.4 & 89.1 & \textbf{90.7} & 65 & 78.3 \\ 
        English & eng\_Latn & 5 & \cmark & Indo-European & Europe 1 & 59.91 & 90 & 92.1 & 92.2 & 91.2 & \textbf{92.5} & 71.8 & 76.6 \\ 
        Esperanto & epo\_Latn & 1 & \cmark & Constructed & Europe 1 & 61.37 & 87.5 & 89.4 & 88.5 & 88.6 & \textbf{90.4} & 58.7 & 70.3 \\ 
        Estonian & est\_Latn & 3 & \cmark & Uralic & Europe 1 & 59.56 & 84.4 & 88.9 & 89 & 90.2 & \textbf{90.4} & 61.9 & 74.4 \\ 
        Basque & eus\_Latn & 4 & \cmark & Basque & Europe 1 & 62.88 & 83.5 & 89.2 & 89.2 & 88.9 & \textbf{89.7} & 55.3 & 53.1 \\ 
        Ewe & ewe\_Latn & 1 & ~ & Atlantic-Congo & Africa & \textbf{71.54} & 49.2 & 56.4 & 32.7 & 31.7 & 33.4 & 20.4 & 12.2 \\ 
        Faroese & fao\_Latn & 1 & ~ & Indo-European & Europe 1 & 71.5 & 80.2 & \textbf{85.3} & 78.1 & 79 & 78.6 & 49.5 & 54.2 \\ 
        Fijian & fij\_Latn & 1 & ~ & Austronesian & Oceania & \textbf{70.56} & 54 & 60.7 & 38.4 & 38 & 38.4 & 40.1 & 39.1 \\ 
        Finnish & fin\_Latn & 4 & \cmark & Uralic & Europe 1 & 67.01 & 89.2 & \textbf{91.6} & 89 & 90.1 & 90.1 & 65 & 74.9 \\ 
        Fon & fon\_Latn & ~ & ~ & Atlantic-Congo & Africa & \textbf{65.94} & 46.2 & 48.1 & 35.4 & 30.6 & 32.7 & 13.8 & 10.8 \\ 
        French & fra\_Latn & 5 & \cmark & Indo-European & Europe 1 & 57.57 & 89.2 & 89.7 & 89.5 & 89.7 & \textbf{90.1} & 73.2 & 77.5 \\ 
        Friulian & fur\_Latn & 1 & ~ & Indo-European & Europe 1 & 63.24 & 77.2 & \textbf{83.2} & 72.9 & 73.9 & 73 & 41.8 & 40.2 \\ 
        Nigerian Fulfulde & fuv\_Latn & 0 & ~ & Atlantic-Congo & Africa & 58.9 & 53.9 & \textbf{63} & 45.6 & 46.1 & 46.6 & 15.5 & 13.4 \\ 
        West Central Oromo & gaz\_Latn & 1 & \cmark & Afro-Asiatic & Africa & 53.21 & 34.2 & \textbf{62} & 45.8 & 48.7 & 43.3 & 27.2 & 18.4 \\ 
        Scottish Gaelic & gla\_Latn & 0 & \cmark & Indo-European & Europe 1 & 59.21 & 60.7 & \textbf{79.9} & 71.7 & 73.9 & 73.2 & 49.7 & 61.5 \\ 
        Irish & gle\_Latn & 2 & \cmark & Indo-European & Europe 1 & 58.59 & 72 & \textbf{84.6} & 81.3 & 82.5 & 82.1 & 52.3 & 71.4 \\ 
        Galician & glg\_Latn & 3 & \cmark & Indo-European & Europe 1 & 67.7 & 89.2 & 88.2 & 89.6 & 90 & \textbf{91.5} & 59.5 & 76.5 \\ 
        Guarani & grn\_Latn & 1 & ~ & Tupian & Americas & 57.71 & 69.6 & \textbf{76.3} & 61.3 & 61.1 & 61.7 & 32.3 & 28.2 \\ 
        Gujarati & guj\_Gujr & 1 & \cmark & Indo-European & Asia 2 & 60.59 & 83.8 & \textbf{87.8} & 87.1 & 87.5 & 86.9 & 65 & 69.2 \\ 
        Haitian Creole & hat\_Latn & 0 & ~ & Indo-European & Americas & 59.21 & 57.3 & \textbf{73.1} & 54 & 55.2 & 54 & 52.2 & 67.2 \\ 
        Hausa & hau\_Latn & 2 & \cmark & Afro-Asiatic & Africa & 55.99 & 72.5 & \textbf{80.9} & 78.3 & 80.4 & 77.2 & 38.2 & 45.1 \\ 
        Hebrew & heb\_Hebr & 3 & \cmark & Afro-Asiatic & Asia 1 & 59.44 & 86.5 & 87.1 & 87.1 & 87.4 & \textbf{88.2} & 60.5 & 73.3 \\ 
        Hindi & hin\_Deva & 4 & \cmark & Indo-European & Asia 2 & 66.71 & 83.9 & \textbf{90.7} & 90.1 & 90.4 & 89.9 & 63.1 & 79.3 \\ 
        Chhattisgarhi & hne\_Deva & ~ & ~ & Indo-European & Asia 2 & 68.17 & 82.8 & \textbf{87.5} & 85.3 & 86.6 & 86 & 52.7 & 63.6 \\ 
        Croatian & hrv\_Latn & 4 & \cmark & Indo-European & Europe 1 & 60.92 & 89.7 & 90.7 & 90.7 & 90.3 & \textbf{91.5} & 66.6 & 77.7 \\ 
        Hungarian & hun\_Latn & 4 & \cmark & Uralic & Europe 1 & 59.86 & 86 & 88.3 & 89.3 & 88.6 & \textbf{90} & 60.3 & 75 \\ 
        Armenian & hye\_Armn & 1 & \cmark & Indo-European & Asia 1 & 65.17 & 86.1 & 88.7 & 88.3 & 88.3 & \textbf{89.2} & 40.1 & 70 \\ 
        Igbo & ibo\_Latn & 1 & ~ & Atlantic-Congo & Africa & \textbf{63.87} & 46.3 & 57.5 & 34.6 & 35.7 & 33.2 & 31.3 & 38.4 \\ 
        Ilocano & ilo\_Latn & 1 & ~ & Austronesian & Asia 3 & 63.11 & 72.3 & \textbf{76.3} & 64.4 & 66.9 & 66.6 & 59.6 & 58.9 \\ 
        Indonesian & ind\_Latn & 3 & \cmark & Austronesian & Asia 3 & 56.54 & 88.9 & 91 & 91.7 & \textbf{92} & 91.6 & 67.3 & 75.1 \\ 
        Icelandic & isl\_Latn & 2 & \cmark & Indo-European & Europe 1 & 62.83 & 85.1 & 89.5 & 90.4 & 90 & \textbf{91.4} & 62.5 & 72.1 \\ 
        Italian & ita\_Latn & 4 & \cmark & Indo-European & Europe 1 & 54.05 & 89 & 90.6 & 90.2 & \textbf{90.8} & 90.6 & 62.8 & 81 \\ 
        Javanese & jav\_Latn & 1 & \cmark & Austronesian & Asia 3 & 57.99 & 81.7 & \textbf{83.2} & 81.8 & 83 & 81.9 & 43 & 50.3 \\ 
        Japanese & jpn\_Jpan & 5 & \cmark & Japonic & Asia 3 & 73.83 & 87.9 & \textbf{89.9} & 89.3 & 88.8 & 89.7 & 63.4 & 75.8 \\ 
        Kabyle & kab\_Latn & 1 & ~ & Afro-Asiatic & Africa & \textbf{61.13} & 36 & 39.5 & 26.3 & 24.5 & 26 & 15.2 & 8 \\ 
        Jingpho & kac\_Latn & 0 & ~ & Sino-Tibetan & Asia 3 & \textbf{64.75} & 54.7 & 62.7 & 35.2 & 33.9 & 35.7 & 10 & 7.8 \\ 
        Kamba & kam\_Latn & 0 & ~ & Atlantic-Congo & Africa & \textbf{56.67} & 48.2 & 52.5 & 39 & 38.4 & 40.1 & 19.7 & 18.7 \\ 
        Kannada & kan\_Knda & 1 & \cmark & Dravidian & Asia 2 & 65.56 & 86.5 & \textbf{90.1} & 89.6 & 89.7 & 89.9 & 60.1 & 69.1 \\ 
        Kashmiri (Arabic) & kas\_Arab & 1 & ~ & Indo-European & Asia 2 & 65.16 & 68 & \textbf{77.4} & 67.8 & 70 & 69.1 & 33 & 46 \\ 
        Kashmiri (Devanagari) & kas\_Deva & 1 & ~ & Indo-European & Asia 2 & 61.49 & 62.8 & \textbf{74.4} & 60.2 & 63.7 & 59.3 & 22.9 & 21.7 \\ 
        Georgian & kat\_Geor & 3 & \cmark & Kartvelian & Asia 1 & 63.72 & 83.4 & 88.5 & 89.1 & 88.6 & \textbf{89.7} & 44.7 & 66.1 \\ 
        Kazakh & kaz\_Cyrl & 3 & \cmark & Turkic & Asia 1 & 63.37 & 85.3 & \textbf{91.4} & 90.9 & 89.7 & 89 & 63.7 & 71.2 \\ 
        Kabiyè & kbp\_Latn & 1 & ~ & Atlantic-Congo & Africa & \textbf{69.88} & 37.9 & 49.1 & 30.4 & 29.8 & 30.5 & 23.7 & 9.3 \\ 
        Kabuverdianu & kea\_Latn & ~ & ~ & Indo-European & Africa & 64.26 & 78.9 & \textbf{86.1} & 73 & 75.7 & 72.3 & 47.5 & 48.8 \\ 
        Halh Mongolian & khk\_Cyrl & 1 & \cmark & Mongolic-Khitan & Asia 3 & 66.21 & 82.9 & \textbf{88.5} & 86.1 & 86.2 & 85.5 & 57.7 & 67.6 \\ 
        Khmer & khm\_Khmr & 1 & \cmark & Austroasiatic & Asia 3 & 74.08 & 85.8 & \textbf{89.2} & 87.5 & 86.8 & 87.7 & 39.2 & 70.4 \\ 
        Kikuyu & kik\_Latn & 1 & ~ & Atlantic-Congo & Africa & \textbf{60.47} & 50 & 59.9 & 38.3 & 39.4 & 38.5 & 28.5 & 28.6 \\ 
        Kinyarwanda & kin\_Latn & 1 & ~ & Atlantic-Congo & Africa & \textbf{68.77} & 45.3 & 48 & 34.5 & 35.1 & 35.7 & 47.1 & 50.5 \\ 
        Kyrgyz & kir\_Cyrl & ~ & \cmark & Turkic & Asia 1 & 58.8 & 84.6 & \textbf{88.1} & 87.9 & 87.1 & 87.5 & 49.8 & 57 \\ 
        Kimbundu & kmb\_Latn & ~ & ~ & Atlantic-Congo & Africa & \textbf{54.4} & 40.9 & 49.7 & 34.6 & 36.4 & 35.7 & 13 & 11.2 \\ 
        Northern Kurdish & kmr\_Latn & 1 & \cmark & Indo-European & Asia 1 & 64.38 & 75.8 & \textbf{81.6} & 77.3 & 79.4 & 78.3 & 42.5 & 46.8 \\ 
        Kanuri (Arabic) & knc\_Arab & 0 & ~ & Nilotic & Africa & \textbf{48.2} & 39 & 41.8 & 23.9 & 21.6 & 21.2 & 2.2 & 1.8 \\ 
        Kanuri (Latin) & knc\_Latn & 0 & ~ & Nilotic & Africa & 60.21 & 58.2 & \textbf{61.8} & 41.4 & 41.8 & 41.3 & 16.8 & 12 \\ 
        Kikongo & kon\_Latn & ~ & ~ & Atlantic-Congo & Africa & 58.59 & 58.5 & \textbf{65} & 44.5 & 46.8 & 46.3 & 18.4 & 23.5 \\ 
        Korean & kor\_Hang & 4 & \cmark & Koreanic & Asia 3 & 67.81 & 86.5 & 88.5 & 88.7 & 88.7 & \textbf{89.2} & 67.8 & 78.2 \\ 
        Lao & lao\_Laoo & 2 & \cmark & Tai-Kadai & Asia 3 & 69.66 & \textbf{88.2} & 87.6 & 85.9 & 86.8 & 86.9 & 26.3 & 44 \\ 
        Ligurian & lij\_Latn & 1 & ~ & Indo-European & Europe 1 & 56.12 & 76.2 & \textbf{81.3} & 75.1 & 75.7 & 74.3 & 29.2 & 43.2 \\ 
        Limburgish & lim\_Latn & 1 & ~ & Indo-European & Europe 1 & 58.8 & 79.1 & \textbf{84.3} & 78.1 & 77.6 & 76.7 & 43.1 & 48.4 \\ 
        Lingala & lin\_Latn & 1 & ~ & Atlantic-Congo & Africa & 61.39 & 60.6 & \textbf{65.8} & 42.6 & 43.8 & 45 & 29.1 & 26 \\ 
        Lithuanian & lit\_Latn & 3 & \cmark & Indo-European & Europe 1 & 64.02 & 87.6 & 88.8 & 88.4 & 89.1 & \textbf{89.6} & 59.2 & 75.3 \\ 
        Lombard & lmo\_Latn & 1 & ~ & Indo-European & Europe 1 & 54.75 & 72.6 & \textbf{80.1} & 70.1 & 71.9 & 71.1 & 36.2 & 43 \\ 
        Latgalian & ltg\_Latn & 1 & ~ & Indo-European & Europe 2 & 66.5 & 79 & \textbf{81.1} & 76 & 76.1 & 75.8 & 29.4 & 38 \\ 
        Luxembourgish & ltz\_Latn & 1 & ~ & Indo-European & Europe 1 & 64.95 & 76.1 & \textbf{82} & 70.6 & 71.9 & 70.9 & 62.4 & 63.8 \\ 
        Luba-Kasai & lua\_Latn & ~ & ~ & Atlantic-Congo & Africa & 53.05 & 52.9 & \textbf{56.3} & 43.5 & 43.6 & 44.5 & 25.6 & 17.9 \\ 
        Ganda & lug\_Latn & ~ & ~ & Atlantic-Congo & Africa & \textbf{49.93} & 41.8 & 45 & 32.5 & 33.8 & 33.4 & 36.8 & 29.6 \\ 
        Luo & luo\_Latn & ~ & ~ & Nilotic & Africa & \textbf{62.3} & 51.1 & 60 & 37.8 & 39.3 & 40.1 & 19.7 & 14.4 \\ 
        Mizo & lus\_Latn & 0 & ~ & Sino-Tibetan & Asia 2 & 58.97 & 68.6 & \textbf{70.6} & 54.2 & 56.4 & 56.4 & 48.7 & 35.9 \\ 
        Standard Latvian & lvs\_Latn & 3 & \cmark & Indo-European & Europe 2 & 72.63 & 89.9 & 90 & 90 & 90.1 & \textbf{91.7} & 63.8 & 75.2 \\ 
        Magahi & mag\_Deva & 0 & ~ & Indo-European & Asia 2 & 65.07 & 82.7 & 86.9 & 86.3 & \textbf{87} & 86.1 & 54.2 & 67.1 \\ 
        Maithili & mai\_Deva & 1 & ~ & Indo-European & Asia 2 & 69.94 & 82.5 & \textbf{89.1} & 85.8 & 87.4 & 87 & 49.5 & 66.7 \\ 
        Malayalam & mal\_Mlym & 1 & \cmark & Dravidian & Asia 2 & 63.89 & 83.3 & 85.5 & 86.7 & \textbf{88} & 86.9 & 59.7 & 71.5 \\ 
        Marathi & mar\_Deva & 2 & \cmark & Indo-European & Asia 2 & 63.77 & 83.1 & \textbf{89.9} & 87.2 & 88.4 & 87.3 & 57.8 & 72.8 \\ 
        Minangkabau (Arabic) & min\_Arab & 1 & ~ & Austronesian & Asia 3 & \textbf{52.7} & 34.9 & 38.1 & 24.6 & 23.5 & 23.8 & 8.1 & 11.2 \\ 
        Minangkabau (Latin) & min\_Latn & 1 & ~ & Austronesian & Asia 3 & 52.68 & 81 & \textbf{85.4} & 78.5 & 79.4 & 79.4 & 40.2 & 36.7 \\ 
        Macedonian & mkd\_Cyrl & 1 & \cmark & Indo-European & Europe 2 & 61.69 & 86.7 & \textbf{89.2} & 88.6 & 88 & 88.8 & 62.2 & 76.9 \\ 
        Maltese & mlt\_Latn & 2 & ~ & Afro-Asiatic & Europe 1 & 68.84 & 69.4 & \textbf{78} & 57.2 & 59.3 & 60 & 63.5 & 73.1 \\ 
        Meitei (Bengali script) & mni\_Beng & ~ & ~ & Sino-Tibetan & Asia 2 & \textbf{65.83} & 37.6 & 48.7 & 32.8 & 31 & 28.5 & 4 & 6.9 \\ 
        Mossi & mos\_Latn & ~ & ~ & Atlantic-Congo & Africa & \textbf{69.58} & 52.2 & 59.5 & 35.1 & 36.9 & 35.3 & 14.3 & 7.5 \\ 
        Maori & mri\_Latn & 1 & ~ & Austronesian & Oceania & \textbf{61.19} & 44 & 53.9 & 32.4 & 32.1 & 32.4 & 45.8 & 60 \\ 
        Burmese & mya\_Mymr & 1 & \cmark & Sino-Tibetan & Asia 3 & 65.73 & 81.1 & \textbf{87.2} & 86.2 & 85.9 & 85.3 & 19 & 62.2 \\ 
        Dutch & nld\_Latn & 4 & \cmark & Indo-European & Europe 1 & 59.26 & 88.1 & \textbf{90.6} & 89.1 & 88.7 & 89.2 & 68.9 & 79.2 \\ 
        N'ko & nqo\_Nkoo &  &  & Mande & Africa & 70.3 & - & 23.2 & 25.9 &25.9 & 25.9 & 4.6 & 3.6 \\ 
        Norwegian Nynorsk & nno\_Latn & 1 & \cmark & Indo-European & Europe 1 & 63.2 & 85.8 & 89.6 & 89.2 & 88.8 & \textbf{90.5} & 62.9 & 75.5 \\ 
        Norwegian Bokmål & nob\_Latn & 1 & \cmark & Indo-European & Europe 1 & 61.04 & 86.4 & \textbf{90} & 88.7 & 88.5 & \textbf{90} & 64.6 & 78.1 \\ 
        Nepali & npi\_Deva & 1 & \cmark & Indo-European & Asia 2 & 69.42 & 85.1 & \textbf{88.5} & 88 & 87.4 & \textbf{88.5} & 62.6 & 68.2 \\ 
        Northern Sotho & nso\_Latn & 1 & ~ & Atlantic-Congo & Africa & \textbf{62.61} & 50.2 & 54.8 & 38.9 & 38.6 & 38.3 & 35.6 & 42.7 \\ 
        Nuer & nus\_Latn & 0 & ~ & Nilotic & Africa & \textbf{57.58} & 41.4 & 43.9 & 31.9 & 27.1 & 29 & 18 & 7.2 \\ 
        Nyanja & nya\_Latn & ~ & ~ & Atlantic-Congo & Africa & \textbf{62.16} & 53 & 60.7 & 46.7 & 46.9 & 49.2 & 46.8 & 45.7 \\ 
        Occitan & oci\_Latn & 1 & ~ & Indo-European & Europe 1 & 60.88 & 82.3 & \textbf{87.5} & 85.7 & 85.8 & 85.6 & 56.8 & 67.5 \\ 
        Odia & ory\_Orya & 1 & \cmark & Indo-European & Asia 2 & 62.83 & 82.1 & \textbf{89.2} & 84.3 & 84.6 & 83.8 & 56.2 & 67.1 \\ 
        Pangasinan & pag\_Latn & 1 & ~ & Austronesian & Asia 3 & 60.17 & 74 & \textbf{78.5} & 68.8 & 71.4 & 70.8 & 54.6 & 44.8 \\ 
        Eastern Panjabi & pan\_Guru & 2 & \cmark & Indo-European & Asia 2 & 66.83 & 82.5 & \textbf{86.3} & 84.4 & 84.9 & 83.2 & 67.6 & 70.9 \\ 
        Papiamento & pap\_Latn & ~ & ~ & Indo-European & Americas & 64.55 & 77.5 & \textbf{84.2} & 72.8 & 74.2 & 71.6 & 56.3 & 61.7 \\ 
        Southern Pashto & pbt\_Arab & 1 & \cmark & Indo-European & Asia 1 & 56.39 & 80.5 & \textbf{81.8} & 80.7 & 81.7 & 80.8 & 52.3 & 55.3 \\ 
        Western Persian & pes\_Arab & 4 & \cmark & Indo-European & Asia 1 & 65.72 & 88.6 & \textbf{91} & 90.4 & 89.7 & 90.2 & 64.3 & 76.4 \\ 
        Plateau Malagasy & plt\_Latn & 1 & \cmark & Austronesian & Africa & 61.09 & 74.1 & \textbf{85.3} & 76.6 & 78.2 & 74.3 & 42.8 & 46.1 \\ 
        Polish & pol\_Latn & 4 & \cmark & Indo-European & Europe 2 & 60.72 & 88 & 90.3 & 89.8 & \textbf{90.7} & \textbf{90.7} & 66.8 & 77.4 \\ 
        Portuguese & por\_Latn & 4 & \cmark & Indo-European & Europe 1 & 62.62 & 89.6 & 89.6 & 88.9 & 88.7 & \textbf{90.1} & 61.7 & 77.8 \\ 
        Dari & prs\_Arab & 0 & ~ & Indo-European & Asia 1 & 64.65 & 86 & 89.1 & 88.6 & 89.2 & \textbf{89.7} & 59.5 & 72.7 \\ 
        Ayacucho Quechua & quy\_Latn & 1 & ~ & Quechuan & Americas & 53.72 & 57.9 & \textbf{64.1} & 46.3 & 50.8 & 49.1 & 36.2 & 18.5 \\ 
        Romanian & ron\_Latn & 3 & \cmark & Indo-European & Europe 2 & 57.25 & 87 & 90.3 & 90.3 & 89.9 & \textbf{91} & 66.5 & 75.1 \\ 
        Rundi & run\_Latn & 0 & ~ & Atlantic-Congo & Africa & \textbf{62.66} & 45.7 & 46 & 35.5 & 37.4 & 39 & 40.8 & 33.1 \\ 
        Russian & rus\_Cyrl & 4 & \cmark & Indo-European & Europe 2 & 64.04 & 88.8 & 89.2 & 88.4 & 88.1 & \textbf{89.7} & 62.8 & 80.2 \\ 
        Sango & sag\_Latn & 1 & ~ & Atlantic-Congo & Africa & \textbf{63.97} & 54.3 & 61 & 42.2 & 42.5 & 43.1 & 15.3 & 9.4 \\ 
        Sanskrit & san\_Deva & 2 & \cmark & Indo-European & Asia 2 & 60.87 & 81.6 & \textbf{85.9} & 80.8 & 82 & 82.5 & 37.8 & 53.3 \\ 
        Santali & sat\_Olck & 1 & ~ & Austroasiatic & Asia 2 & \textbf{55.63} & 24.5 & 24.6 & 24.5 & 21.9 & 22.7 & 1.8 & 0 \\ 
        Sicilian & scn\_Latn & 1 & ~ & Indo-European & Europe 1 & 59.22 & 75.8 & \textbf{81.1} & 68.1 & 72.1 & 70.6 & 43.9 & 58.1 \\ 
        Shan & shn\_Mymr & 0 & ~ & Tai-Kadai & Asia 3 & \textbf{63.92} & 36.9 & 51.8 & 29.7 & 31 & 28.9 & 19.3 & 14.7 \\ 
        Sinhala & sin\_Sinh & 0 & \cmark & Indo-European & Asia 2 & 64.79 & 85.8 & \textbf{87.6} & 86.2 & 87.4 & 87.2 & 22.5 & 70.3 \\ 
        Slovak & slk\_Latn & 3 & \cmark & Indo-European & Europe 2 & 63.52 & 87.1 & \textbf{89.6} & 89.2 & 88.7 & 88.8 & 62.6 & 71.7 \\ 
        Slovenian & slv\_Latn & 3 & \cmark & Indo-European & Europe 1 & 62.71 & 89.7 & \textbf{90.6} & 90.1 & 89.8 & 90 & 61.5 & 72.3 \\ 
        Samoan & smo\_Latn & 1 & ~ & Austronesian & Oceania & \textbf{62.93} & 50.8 & 59.1 & 37.5 & 36 & 37.5 & 50.8 & 54.3 \\ 
        Shona & sna\_Latn & 1 & ~ & Atlantic-Congo & Africa & \textbf{60.97} & 46.7 & 51.5 & 39.2 & 41 & 40.6 & 37.2 & 37.2 \\ 
        Sindhi & snd\_Arab & 1 & \cmark & Indo-European & Asia 2 & 59.71 & 75.4 & \textbf{85.1} & 84.3 & 83.8 & 83.9 & 49.1 & 62.4 \\ 
        Somali & som\_Latn & 1 & \cmark & Afro-Asiatic & Africa & 61.2 & 65.1 & \textbf{78.7} & 72.5 & 74 & 72.8 & 48 & 58.3 \\ 
        Southern Sotho & sot\_Latn & 0 & ~ & Atlantic-Congo & Africa & \textbf{67.74} & 51.6 & 55.9 & 35.9 & 37.3 & 36.8 & 38.2 & 43.2 \\ 
        Spanish & spa\_Latn & 5 & \cmark & Indo-European & Europe 1 & 58.28 & 87.4 & 88.6 & \textbf{89.1} & 88 & 87.9 & 67.5 & 80.6 \\ 
        Sardinian & srd\_Latn & 1 & ~ & Indo-European & Europe 1 & 59.81 & 74.3 & \textbf{80.5} & 67.7 & 71.6 & 69.7 & 36.1 & 50.6 \\ 
        Serbian & srp\_Cyrl & 4 & \cmark & Indo-European & Europe 2 & 58.19 & 89.2 & 89.3 & \textbf{90.1} & \textbf{90.1} & \textbf{90.1} & 55.4 & 75.9 \\ 
        Swati & ssw\_Latn & 1 & ~ & Atlantic-Congo & Africa & 58.03 & 41.9 & \textbf{59} & 45.4 & 46 & 43.6 & 26.1 & 30.3 \\ 
        Sundanese & sun\_Latn & 1 & \cmark & Austronesian & Asia 3 & 51.97 & 80.3 & \textbf{86.9} & 85.2 & 85.9 & 85.6 & 50.8 & 56.7 \\ 
        Swedish & swe\_Latn & 4 & \cmark & Indo-European & Europe 1 & 61 & 88.3 & 90.6 & 90.1 & 89.8 & \textbf{90.8} & 66 & 75.5 \\ 
        Swahili & swh\_Latn & 2 & \cmark & Atlantic-Congo & Africa & 58.91 & 77 & \textbf{85.8} & 81.7 & 83.7 & 82.8 & 63.8 & 71.6 \\ 
        Silesian & szl\_Latn & 1 & ~ & Indo-European & Europe 2 & 57.48 & 79.4 & \textbf{84.9} & 79.1 & 79.9 & 79 & 38.4 & 51.7 \\ 
        Tamil & tam\_Taml & 3 & \cmark & Dravidian & Asia 2 & 66.53 & 86.9 & 87 & 87.6 & 87.6 & \textbf{88.8} & 48.7 & 69.2 \\ 
        Tamasheq (Latin) & taq\_Latn & ~ & ~ & Afro-Asiatic & Africa & 52.82 & 52.9 & \textbf{55.1} & 38.1 & 38.4 & 37.9 & 22.8 & 17.7 \\ 
        Tamasheq (Tifinagh) & taq\_Tfng & ~ & ~ & Afro-Asiatic & Africa & \textbf{52.1} & 27.1 & 26.9 & 28 & 28 & 27.8 & 3.1 & 3.3 \\ 
        Tatar & tat\_Cyrl & 1 & ~ & Turkic & Europe 2 & 65.64 & 75 & \textbf{81.9} & 69.8 & 71.6 & 71.1 & 46.1 & 61.9 \\ 
        Telugu & tel\_Telu & 1 & \cmark & Dravidian & Asia 2 & 62.79 & 81 & \textbf{88.8} & 87.3 & 86.8 & 87.1 & 60.4 & 68.4 \\ 
        Tajik & tgk\_Cyrl & 1 & ~ & Indo-European & Asia 1 & \textbf{66.98} & 46.9 & 59.8 & 37.5 & 38.8 & 37.9 & 49.1 & 61.6 \\ 
        Tagalog & tgl\_Latn & 3 & ~ & Austronesian & Asia 3 & 55.19 & 84.7 & 86.2 & 85.6 & \textbf{86.8} & 86.7 & 67.9 & 76.9 \\ 
        Thai & tha\_Thai & 3 & \cmark & Tai-Kadai & Asia 3 & 69.47 & 87.5 & \textbf{91} & 89.7 & 89.7 & 90.8 & 61.1 & 75.5 \\ 
        Tigrinya & tir\_Ethi & 2 & ~ & Afro-Asiatic & Africa & 57.04 & 53.9 & \textbf{67.7} & 49.1 & 52.9 & 50 & 19.3 & 37.1 \\ 
        Tok Pisin & tpi\_Latn & 1 & ~ & Indo-European & Oceania & 69.99 & 75.9 & \textbf{79.4} & 66.1 & 67.2 & 66.5 & 63.8 & 60.8 \\ 
        Tswana & tsn\_Latn & 2 & ~ & Atlantic-Congo & Africa & \textbf{61.92} & 47.8 & 58.5 & 34.6 & 33.7 & 35.6 & 31.8 & 42.8 \\ 
        Tsonga & tso\_Latn & 1 & ~ & Atlantic-Congo & Africa & \textbf{63.8} & 52.7 & 57.2 & 35.9 & 38 & 38.8 & 34.2 & 29.4 \\ 
        Turkmen & tuk\_Latn & 1 & ~ & Turkic & Asia 1 & 65.9 & 64.4 & \textbf{76.8} & 60.3 & 61.2 & 59.6 & 52.3 & 53.1 \\ 
        Tumbuka & tum\_Latn & 1 & ~ & Atlantic-Congo & Africa & \textbf{75.92} & 53.8 & 66 & 36.1 & 39.9 & 38.1 & 28.5 & 32.6 \\ 
        Turkish & tur\_Latn & 4 & \cmark & Turkic & Asia 1 & 65.4 & 87.4 & 90.4 & 89.8 & 89.9 & \textbf{91.1} & 66.1 & 78.2 \\ 
        Twi & twi\_Latn & 1 & ~ & Atlantic-Congo & Africa & \textbf{63.43} & 57.4 & 62.2 & 44 & 42.6 & 43.8 & 33 & 34.6 \\ 
        Tamazight & tzm\_Tfng & ~ & ~ & Afro-Asiatic & Africa & \textbf{61.18} & 24.4 & 23.9 & 25.5 & 25.2 & 24.3 & 4.1 & 1.1 \\ 
        Uyghur & uig\_Arab & 1 & \cmark & Turkic & Asia 1 & 69.58 & 81.5 & 85.8 & 83.9 & \textbf{85.9} & 83.9 & 46.4 & 57.7 \\ 
        Ukrainian & ukr\_Cyrl & 3 & \cmark & Indo-European & Europe 2 & 61.68 & 89.6 & 91.6 & 91.7 & 90.6 & \textbf{92.2} & 67.1 & 79.2 \\ 
        Umbundu & umb\_Latn & 0 & ~ & Atlantic-Congo & Africa & \textbf{54.93} & 48.8 & 53.6 & 35.8 & 36.8 & 35 & 15.2 & 7.7 \\ 
        Urdu & urd\_Arab & 3 & \cmark & Indo-European & Asia 2 & 61.51 & 84.3 & \textbf{87.6} & 86.2 & 86.5 & 85.8 & 64.3 & 76.8 \\ 
        Northern Uzbek & uzn\_Latn & 3 & \cmark & Turkic & Asia 1 & 59.2 & 82.1 & 85.7 & 87.5 & \textbf{87.6} & 87.3 & 57.6 & 60.3 \\ 
        Venetian & vec\_Latn & 1 & ~ & Indo-European & Europe 1 & 52.64 & 79.6 & \textbf{84.3} & 79.3 & 79.7 & 78.9 & 44.1 & 59.4 \\ 
        Vietnamese & vie\_Latn & 4 & \cmark & Austroasiatic & Asia 3 & 69.73 & 88.3 & 90.5 & 90.4 & 89.9 & \textbf{91.2} & 63.3 & 75.7 \\ 
        Waray & war\_Latn & 1 & ~ & Austronesian & Asia 3 & 62.36 & 77.9 & \textbf{80.7} & 75.1 & 78.7 & 77.5 & 64.6 & 65.4 \\ 
        Wolof & wol\_Latn & 2 & ~ & Atlantic-Congo & Africa & 52.67 & 57.1 & \textbf{60.1} & 46.1 & 46.8 & 47.9 & 29.5 & 24.1 \\ 
        Xhosa & xho\_Latn & 2 & \cmark & Atlantic-Congo & Africa & 62.29 & 54 & \textbf{70.7} & 62.3 & 64.3 & 60.7 & 37.7 & 48.2 \\ 
        Eastern Yiddish & ydd\_Hebr & 1 & \cmark & Indo-European & Europe 1 & 52.3 & 57.5 & \textbf{82.9} & 72.2 & 74.4 & 70.5 & 33.2 & 44.8 \\ 
        Yoruba & yor\_Latn & 2 & ~ & Atlantic-Congo & Africa & \textbf{62.55} & 42.4 & 49.6 & 33.6 & 32.5 & 30.6 & 33.8 & 40.5 \\ 
        Yue Chinese & yue\_Hant & ~ & ~ & Sino-Tibetan & Asia 3 & 70.52 & 88.6 & \textbf{88.9} & 88.2 & 88.1 & 88 & 63.9 & 79.2 \\ 
        Chinese (Simplified) & zho\_Hans & 5 & \cmark & Sino-Tibetan & Asia 3 & 74.67 & \textbf{90.2} & 90.1 & 89.1 & 88.8 & 89.9 & 60.6 & 79.5 \\ 
        Chinese (Traditional) & zho\_Hant & 1 & \cmark & Sino-Tibetan & Asia 3 & 75.2 & 88.2 & \textbf{91.6} & 90.1 & 89.1 & 89.6 & 64.2 & 78.8 \\ 
        Standard Malay & zsm\_Latn & 3 & \cmark & Austronesian & Asia 3 & 61.39 & 90 & \textbf{92.7} & 91.3 & 91.5 & 91.3 & 63.4 & 76.1 \\ 
        Zulu & zul\_Latn & 2 & ~ & Atlantic-Congo & Africa & 60.13 & 48.4 & \textbf{73.5} & 62 & 63.6 & 60.1 & 39.3 & 53.1 \\ 
        \midrule
        \textbf{Average} &  & ~ & ~ & ~ & ~ & 62.3 & 70.9 & \textbf{76.1} & 68.8 & 69.3 & 69.1 & 45.1 & 52.6 \\ 
        \bottomrule
    \caption{{\bf Overall result of the performance of different text classification models across different languages.} We compared different settings: fully-supervised, cross-lingual transfer and zero-shot prompting of LLMs. We report cross-lingual transfer performances from 4 source languages: English, French, Chinese and Arabic.}\\
     \label{tab:all_language_results}
    \end{longtable}
\end{scriptsize}

\begin{table}[!ht]
    \centering
    \setlength\tabcolsep{4pt}
    \scalebox{0.6}{
    \begin{tabular}{llccccc|rrrrrr}
      & \textbf{Language} & \textbf{in} & \textbf{in Afri-}& \textbf{in Afro-} & \textbf{in Afro} & \textbf{in Afro} &  & \textbf{Afri} & \textbf{Afro} & \textbf{Afro} & \textbf{Afro} & \textbf{Afro} \\ 
    \textbf{Language} & \textbf{code} & \textbf{XLM-R?} & \textbf{BERTa?} & \textbf{XLMR?} & \textbf{XLMR-61} & \textbf{XLMR-76} & \textbf{XLMR} & \textbf{BERTa} & \textbf{XLMR} & \textbf{XLMR-61} & \textbf{XLM-76} & \textbf{XLM-76-script} \\ 
    \midrule
        Tunisian Arabic & aeb\_Arab & \cmark & ~ & \cmark & \cmark & \cmark & 86.5 & 25.4 & 86.1 & 86.7 & 86.8 & \textbf{87.0} \\ 
        Moroccan Arabic & ary\_Arab & \cmark & ~ & \cmark & \cmark & \cmark & \textbf{90.1} & 26 & 87.2 & 87.3 & 88.0  & 88.3\\ 
        Egyptian Arabic & arz\_Arab & \cmark & ~ & \cmark & \cmark & \cmark & 89.1 & 26.4 & 88.7 & 86.6 & 88.8  & \textbf{89.9}\\ 
        Afrikaans & afr\_Latn & \cmark & ~ & \cmark & \cmark & \cmark & 89.8 & 53.7 & 90.4 & 89.1 & \textbf{91.1} & 90.0\\ 
        Akan & aka\_Latn & ~ & ~ & ~ & \cmark & \cmark & 59.7 & 52.6 & 59.4 & 74.9 & \textbf{79.8} & 76.9 \\ 
        Amharic & amh\_Ethi & \cmark & \cmark & \cmark & \cmark & \cmark & 84.2 & 80.2 & \textbf{88.6} & 87.2 & 87.1 & 86.4\\ 
        Bambara & bam\_Latn & ~ & ~ & ~ & ~ & \cmark & 49.3 & 55.4 & 59.3 & 59.4 & 70.9 & \textbf{72.2} \\ 
        Bemba & bem\_Latn & ~ & ~ & ~ & \cmark & \cmark & 59.5 & 55.7 & 74.1 & \textbf{80.8} & 73.6 & 80.4\\ 
        Chokwe & cjk\_Latn & ~ & ~ & ~ & ~ & \cmark & 47.5 & 40.9 & 48.9 & 56.6 & \textbf{63.5} & 60.1 \\ 
        Dinka & dik\_Latn & ~ & ~ & ~ & ~ & \cmark & 61 & 62.3 & 60.9 & 61.3 & 66.4 & \textbf{67.9} \\ 
        Dyula & dyu\_Latn & ~ & ~ & ~ & ~ & \cmark & 48 & 43.8 & 54.9 & 52.6 & \textbf{57.3} & \textbf{57.3}\\ 
        Ewe & ewe\_Latn & ~ & ~ & ~ & \cmark & \cmark & 56.4 & 61.6 & 59.5 & 71.4 & \textbf{78.7} & 77.8 \\ 
        Fon & fon\_Latn & ~ & ~ & ~ & \cmark & \cmark & 48.1 & 54.7 & 54.5 & 61.7 & \textbf{68.5} & 67.5 \\ 
        Nigerian Fulfulde & fuv\_Latn & ~ & ~ & ~ & \cmark & \cmark & 63 & 57.5 & 60.8 & 62.5 & \textbf{70.0} & 67.1 \\ 
        Oromo & gaz\_Latn & \cmark & \cmark & \cmark & \cmark & \cmark & 62 & 74.6 & \textbf{82.6} & 81.2 & 77.5 & 75.4 \\ 
        Hausa & hau\_Latn & \cmark & \cmark & \cmark & \cmark & \cmark & 80.9 & 80.4 & \textbf{86.4} & 85.6 & 84.8 & 85.4 \\ 
        Igbo & ibo\_Latn & ~ & \cmark & \cmark & \cmark & \cmark & 57.5 & 79.6 & 83.7 & \textbf{83.9} & 82.5 & 78.6\\ 
        Kabyle & kab\_Latn & ~ & ~ & ~ & ~ & \cmark & 39.5 & 44 & 35.1 & 34.9 & \textbf{53.0} & 47.0 \\ 
        Kamba & kam\_Latn & ~ & ~ & ~ & ~ & \cmark & 52.5 & 53.7 & 59.6 & 59.4 & 67.7 & \textbf{68.1}\\ 
        Kabiyè & kbp\_Latn & ~ & ~ & ~ & ~ & \cmark & 49.1 & 55.2 & 58.8 & 59.2 & \textbf{70.7} & 70.2 \\ 
        Kanuri (Arabic) & knc\_Arab & ~ & ~ & ~ & ~ & \cmark & 41.8 & 36.1 & 47.7 & \textbf{48.4} & 46.4 & 44.6 \\ 
        Kanuri (Latin) & knc\_Latn & ~ & ~ & ~ & ~ & \cmark & 61.8 & 58.2 & 61.4 & 62.7 & 63.0 & \textbf{63.1} \\ 
        Kikuyu & kik\_Latn & ~ & ~ & ~ & \cmark & \cmark & 59.9 & 57.3 & 65.8 & 71.6 & 80.3 & \textbf{80.8} \\ 
        Kinyarwanda & kin\_Latn & ~ & \cmark & \cmark & \cmark & \cmark & 48 & 79.9 & 84.2 & \textbf{85.3} & 86.6 & 84.2 \\ 
        Kimbundu & kmb\_Latn & ~ & ~ & ~ & ~ & \cmark & 49.7 & 49.9 & 58.5 & 60.3 & \textbf{66.6} & 64.7\\ 
        Kikongo & kon\_Latn & ~ & ~ & ~ & ~ & \cmark & 65 & 61.8 & 70.3 & 74.2 & \textbf{82.0} & 80.0 \\ 
        Lingala & lin\_Latn & ~ & ~ & ~ & \cmark & \cmark & 65.8 & 63.2 & 73.8 & 83.3 & \textbf{86.4} & 85.0 \\ 
        Luba-Kasai & lua\_Latn & ~ & ~ & ~ & \cmark & \cmark & 56.3 & 52 & 65.2 & 70.9 & 73.1 & \textbf{76.8} \\ 
        Ganda & lug\_Latn & ~ & ~ & ~ & \cmark & \cmark & 45 & 46.8 & 61.2 & 67.7 & \textbf{73.8} & 71.6 \\ 
        Luo & luo\_Latn & ~ & ~ & ~ & \cmark & \cmark & 60 & 59.5 & 61.2 & 67.4 & \textbf{77.8} & 77.1 \\ 
        Mossi & mos\_Latn & ~ & ~ & ~ & \cmark & \cmark & 59.5 & 52.1 & 61.9 & 63.8 & \textbf{71,1} & 69.7 \\ 
        N'ko & nqo\_Nkoo & ~ & ~ & ~ &  & \cmark & 23.2 & 22.2 & 22.7 & 22.5 & 22.0 & \textbf{40.5}  \\ 
        Northern Sotho & nso\_Latn & ~ & ~ & ~ & \cmark & \cmark & 54.8 & 51.8 & 80.7 & 82.6 & \textbf{83.3} & 82.4 \\ 
        Nuer & nus\_Latn & ~ & ~ & ~ & ~ & \cmark & 43.9 & 54.7 & 47.5 & 46.2 & \textbf{64.5} & 63.4 \\ 
        Nyanja & nya\_Latn & ~ & ~ & \cmark & \cmark & \cmark & 60.7 & 58.1 & 83.3 & \textbf{86.3} & 83.9 & 83.3 \\ 
        Plateau Malagasy & plt\_Latn & \cmark & ~ & \cmark & \cmark & \cmark & 85.3 & 50.5 & 88.4 & 88.2 & 88.1 & \textbf{89.8}\\ 
        Rundi & run\_Latn & ~ & \cmark & \cmark & \cmark & \cmark & 46 & 77.9 & 82.4 & 83.5 & 83.5 & \textbf{83.9} \\ 
        Sango & sag\_Latn & ~ & ~ & ~ & ~ & \cmark & 61 & 61.4 & 62.1 & 65.4 & 66.5 & \textbf{76.7} \\ 
        Shona & sna\_Latn & ~ & ~ & \cmark & \cmark & \cmark & 51.5 & 55.2 & 81.3 & 80.3 & \textbf{82.8} & 82.0\\ 
        Somali & som\_Latn & \cmark & \cmark & \cmark & \cmark & \cmark & 78.7 & 77.7 & 81.7 & 80 & 80.8 & \textbf{82.0}  \\ 
        Southern Sotho & sot\_Latn & ~ & ~ & \cmark & \cmark & \cmark & 55.9 & 57.4 & 83.7 & \textbf{84.0} & 83.5 & 80.8 \\ 
        Swati & ssw\_Latn & ~ & ~ & ~ & \cmark & \cmark & 59 & 53.5 & 80.6 & \textbf{81.8} & 81.3 & 80.1 \\ 
        Swahili & swh\_Latn & \cmark & \cmark & \cmark & \cmark & \cmark & 85.8 & 85.8 & 87.9 & 87.2 & \textbf{88.5} & 87.4 \\ 
        Tamasheq (Latin) & taq\_Latn & ~ & ~ & ~ & ~ & \cmark & 55.1 & 53.4 & 58.1 & 57.7 & \textbf{60.3} & 58.4 \\ 
        Tamasheq (Tifinagh) & taq\_Tfng & ~ & ~ & ~ & ~ & \cmark & 26.9 & 26.4 & 27.9 & 26 & 25.7 & \textbf{36.1}\\ 
        Tigrinya & tir\_Ethi & ~ & \cmark & ~ & \cmark & \cmark & 67.7 & 70.3 & \textbf{81.5} & 81 & 79.8 & 78.0\\ 
        Tswana & tsn\_Latn & ~ & ~ & ~ & \cmark & \cmark & 58.5 & 58 & 79.4 & \textbf{82.2} & 81.9 & 79.8\\ 
        Tsonga & tso\_Latn & ~ & ~ & ~ & \cmark & \cmark & 57.2 & 58.8 & 68.5 & \textbf{80.9} & 82.5 & 84.2 \\ 
        Tumbuka & tum\_Latn & ~ & ~ & ~ & \cmark & \cmark & 66 & 67.3 & 82.7 & 87.2 & \textbf{87.6} & 86.3 \\ 
        Twi & twi\_Latn & ~ & ~ & ~ & \cmark & \cmark & 62.2 & 64.1 & 65.8 & 77.5 & \textbf{80.2} & 77.3 \\ 
        Tamazight & tzm\_Tfng & ~ & ~ & ~ & ~ & \cmark & 23.9 & \textbf{26.5} & 25.9 & 25 & 26.7 & 55.5\\ 
        Umbundu & umb\_Latn & ~ & ~ & ~ & \cmark & \cmark & 53.6 & 51.8 & 59.9 & 62.3 & \textbf{68.5} & 64.9 \\ 
        Wolof & wol\_Latn & ~ & ~ & ~ & \cmark & \cmark & 60.1 & 50.1 & 64.3 & 66.6 & \textbf{73.3} & 71.7 \\ 
        Xhosa & xho\_Latn & \cmark & ~ & \cmark & \cmark & \cmark & 70.7 & 47.5 & 83.1 & 83.5 & \textbf{84.0} & 82.8\\ 
        Yoruba & yor\_Latn & ~ & ~ & \cmark & \cmark & \cmark & 49.6 & 70.6 & 74.8 & \textbf{80.5} & 78.8 & 76.0 \\ 
        Zulu & zul\_Latn & ~ & ~ & \cmark & \cmark & \cmark & 73.5 & 53.6 & 84.9 & 84.2 & 85.8 & \textbf{86.6} \\ 
        \midrule
        \textbf{Average} &  & ~ & ~ & ~ & ~ & ~ & \textbf{59.9} & \textbf{56.1} & \textbf{69.2} & \textbf{71.6} & \textbf{74.1} & \textbf{74.3} \\ 
        \bottomrule
    \end{tabular}
     }
    \caption{{\bf Evaluation result on different African languages pre-trained language models}}
    \label{tab:african_result_table}
   
\end{table}

\end{document}